\documentclass[acmtog, authorversion, nonacm]{acmart}

\usepackage{booktabs} %

\usepackage[ruled]{algorithm2e} %

\SetAlFnt{\small}
\SetAlCapFnt{\small}
\SetAlCapNameFnt{\small}
\SetAlCapHSkip{0pt}

\usepackage{amsmath,amsfonts,bm}

\def\figref#1{figure~\ref{#1}}
\def\Figref#1{Figure~\ref{#1}}

\def\secref#1{section~\ref{#1}}
\def\Secref#1{Section~\ref{#1}}

\def\eqref#1{equation~\ref{#1}}

\def\1{\bm{1}}

\def\ve{{\bm{e}}}

\def\vi{{\bm{i}}}

\def\vo{{\bm{o}}}

\def\vw{{\bm{w}}}

\def\mW{{\bm{W}}}

\DeclareMathAlphabet{\mathsfit}{\encodingdefault}{\sfdefault}{m}{sl}
\SetMathAlphabet{\mathsfit}{bold}{\encodingdefault}{\sfdefault}{bx}{n}

\newcommand{\R}{\mathbb{R}}

\usepackage{amsmath}
\usepackage{comment}
\usepackage{multirow,bigdelim}
\usepackage{lipsum}
\usepackage{array}
\usepackage{wrapfig}

\usepackage{adjustbox}
\usepackage[percent]{overpic}
\usepackage{makecell}
\usepackage[normalem]{ulem}
\usepackage{blindtext}
\usepackage{xcolor}
\usepackage{soul}
\usepackage{cleveref}
\usepackage{amsfonts}
\usepackage{xspace} %
\usepackage{braket}
\usepackage{nicefrac}
\usepackage[percent]{overpic}
\usepackage{enumitem} %
\usepackage{multicol} %
\usepackage{subfig} 

\makeatletter
\newcommand{\settitle}{\@maketitle}
\makeatother

\newcolumntype{C}[1]{>{\centering\let\newline\\\arraybackslash\hspace{0pt}}m{#1}}

\newif\ifdraft
\drafttrue

\newcommand{\tildeapprox}{{\raise.17ex\hbox{$\scriptstyle\sim$}}}
\renewcommand{\secref}[1]{\Secref{#1}}
\renewcommand{\figref}[1]{\Figref{#1}}

\renewcommand{\eqref}[1]{Eq.~(\ref{#1})}

\definecolor{darkpink}{rgb}{0.561, 0.282, 0.427}
\definecolor{atomictangerine}{rgb}{0.8, 0.2, 0.1}
\definecolor{turq}{rgb}{0.0, 0.5, 0.5}
\definecolor{darkturq}{rgb}{0.0, 0.4, 0.4}
\definecolor{bright}{rgb}{0.8, 0.1, 0}
\definecolor{darkgray}{gray}{0.3}
\definecolor{gray}{gray}{0.5}
\definecolor{mahogany}{rgb}{0.6, 0.05, 0.05}
\definecolor{editblue}{rgb}{0.3,0.05,0.9}
\definecolor{black}{rgb}{0.,0.,0.}
\definecolor{darkgreen}{rgb}{0.1,0.5,0.0}
\definecolor{olive}{rgb}{0.537, 0.627, 0.318}
\definecolor{green}{rgb}{0.22, 0.463, 0.114}
\definecolor{grey}{rgb}{0.4, 0.4, 0.4}
\definecolor{blue}{rgb}{0.435, 0.659, 0.863}
\definecolor{pink}{rgb}{0.761, 0.482, 0.627}
\definecolor{darkpink}{rgb}{0.561, 0.282, 0.427}

\ifdraft

\newcommand\edit[1]{\textcolor{black}{ #1\ignorespaces}}

\newcommand\gal[1]{}
\newcommand\yuval[1]{}
\newcommand\rgc[1]{}
\newcommand\yoad[1]{}

\newcommand{\drop}[1]{}

\else
\newcommand{\dcc}[1]{}
\newcommand{\rgc}[1]{}
\newcommand{\opc}[1]{}
\newcommand{\gcc}[1]{}
\newcommand{\hmc}[1]{}
\newcommand{\abc}[1]{}

\newcommand\gal[1]{}
\newcommand\yuval[1]{}
\newcommand\yoad[1]{}
\fi

\newcommand{\ourmethod}{\textit{Perfusion}\xspace}
\newcommand{\TTI}{T2I\xspace}
\newcommand{\e}{\ve\xspace}
\newcommand{\istar}{\vi_*\xspace}
\newcommand{\ostar}{\vo_*\xspace}
\newcommand{\myparagraph}[1]{\textbf{#1\,}}

\newcommand{\Em}{\ensuremath{\e_{m}}\xspace}  %
\newcommand{\Where}{\textit{Where}\xspace}
\newcommand{\What}{\textit{What}\xspace}

\def\naive{na\"{\i}ve\xspace}
\def\Naive{Na\"{\i}ve\xspace}

\makeatletter
\DeclareRobustCommand\onedot{\futurelet\@let@token\@onedot}
\def\@onedot{\ifx\@let@token.\else.\null\fi\xspace}

\def\eg{\emph{e.g}\onedot}

\def\ie{\emph{i.e}\onedot}

\makeatother

\usepackage[bottom]{footmisc}
\usepackage{arydshln}

\makeatletter
\def\blfootnote{\xdef\@thefnmark{}\@footnotetext}
\makeatother

\begin{document}
\title{Key-Locked Rank One Editing for Text-to-Image Personalization}

\author{Yoad Tewel}
\affiliation{%
 \institution{NVIDIA, Tel-Aviv University}
 \country{Israel}}
\email{tbd@email.com}

\author{Rinon Gal}
\affiliation{%
 \institution{NVIDIA, Tel-Aviv University}
 \country{Israel}}
\email{tbd@email.com}

\author{Gal Chechik}
\affiliation{%
 \institution{NVIDIA, Bar-Ilan University}
 \country{Israel}}
\email{tbd@email.com}

\author{Yuval Atzmon}
\affiliation{%
 \institution{NVIDIA} 
 \country{Israel}}
\email{tbd@email.com}

\begin{abstract}
Text-to-image models (\TTI) offer a new level of flexibility by allowing users to guide the creative process through natural language. However, personalizing these models to align with user-provided visual concepts remains a challenging problem. The task of \TTI personalization poses multiple hard challenges, such as maintaining high visual fidelity while allowing creative control, combining multiple personalized concepts in a single image, and keeping a small model size.

We present \textit{\ourmethod}, a \TTI personalization method that addresses these challenges using dynamic rank-1 updates to the underlying \TTI model.
\ourmethod avoids overfitting by introducing a new mechanism that ``locks'' new concepts' cross-attention Keys to their superordinate category.
Additionally, we develop a \textit{gated} rank-1 approach that enables us to control the influence of a learned concept during inference time and to combine multiple concepts.
This allows runtime-efficient balancing of visual-fidelity and textual-alignment with a single 100KB trained model, which is five orders of magnitude smaller than the current state of the art. Moreover, it can span different operating points across the Pareto front without additional training.

Finally, we show that \textit{\ourmethod} outperforms strong baselines in both qualitative and quantitative terms. Importantly, key-locking leads to novel results compared to traditional approaches, allowing to portray personalized object interactions in unprecedented ways, even in one-shot settings.

Code will be available at our \href{http://research.nvidia.com/labs/par/Perfusion/}{\textcolor{blue}{project page}}\footnote{Accepted to SIGGRAPH 2023}.
\end{abstract}

\begin{teaserfigure}
\centering
\includegraphics[width=0.95\textwidth, trim={0.5cm 26cm 10.9cm 0.02cm},clip]{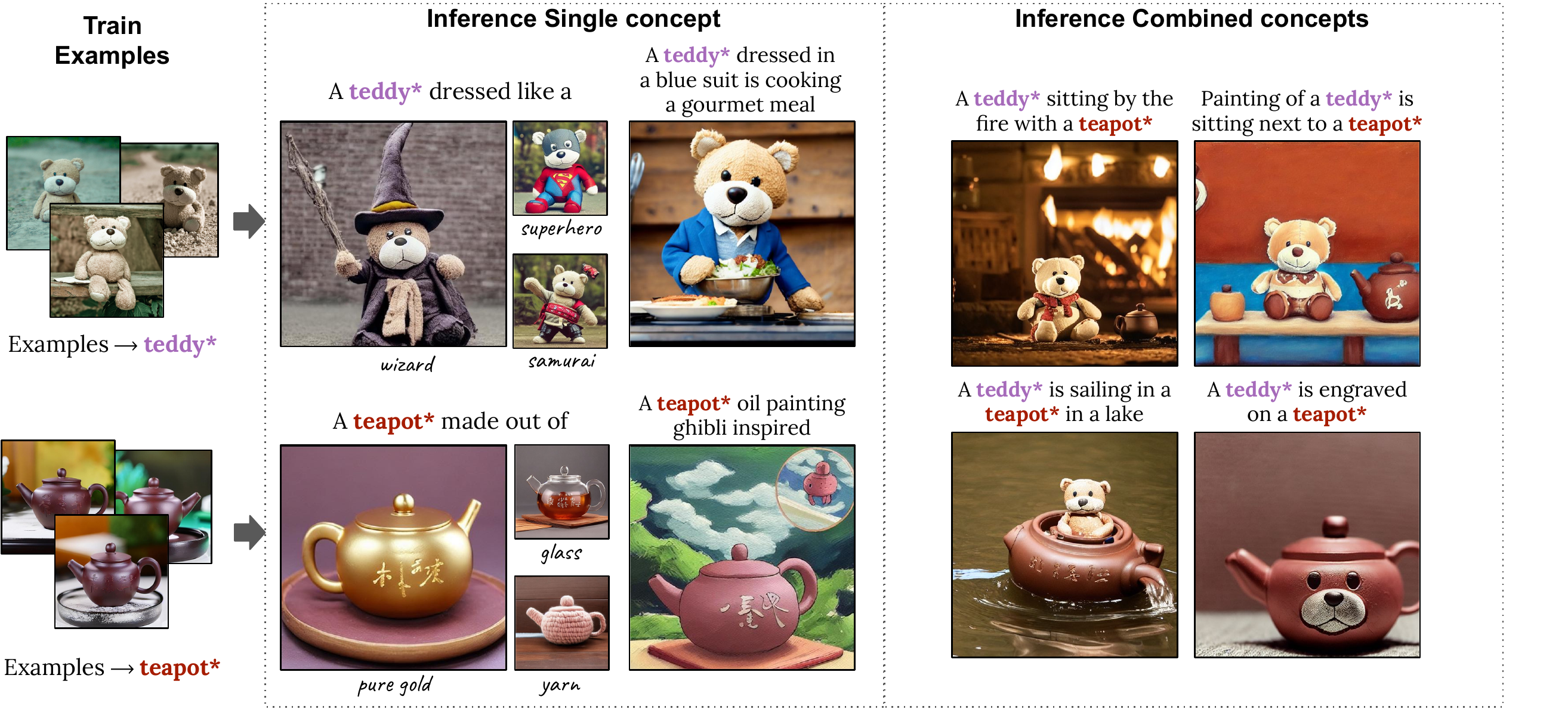} %
    \caption{ (left) \ourmethod enables image
    generation for personalized concepts with large changes in their appearance, pose, and context, using a compact model of only 100KB per concept, without compromising identity. (right) \ourmethod can combine learned concepts at inference time, creating scenes which portray multiple concepts side-by-side, or even create interactions between them. }
    \label{fig1}
\end{teaserfigure}

\maketitle

\section{Introduction}
Text-to-image (\TTI) personalization is the task of customizing a diffusion-based T2I model to reason over novel, user-provided visual concepts~\citep{gal2022textual, ruiz2022dreambooth, cohen2022my, kumari2022customdiffusion}. A user first provides a handful of image examples of the concept; then, they can then use free text to craft novel scenes containing these concepts. This workflow can be used in a wide range of downstream applications from virtual photo shoots through product design to generation of personalized virtual assets.

Current methods for personalization take one of two main approaches. They either represent a concept through a word embedding at the input of the text encoder \cite{gal2022textual, cohen2022my} or fine-tune the full weights of the diffusion-based denoiser module \cite{ruiz2022dreambooth}.
Unfortunately, these approaches
are prone to different types of overfitting. As we show below, word embedding methods struggle to generalize to unseen text prompts. This is reflected in their textual-alignment scores which tend to be low.
Fine-tuning methods can better generalize to new text prompts, but they still lack expressivity, as reflected in their textual and visual alignment scores which tend to be lower than our method. \edit{Moreover, tuning methods typically demand significant storage space, often in the range of hundreds of megabytes or even gigabytes. %
Lastly, both approaches struggle to combine concepts that were trained individually, such as a teddy* and a teapot* (Fig. \ref{fig1}), in a single prompt. }

\vspace{-10pt}
Here we describe ``\textit{\ourmethod}'' a \TTI personalization method aimed \edit{at answering} all these challenges. It allows for \edit{expressive deformations of} the concept while maintaining high concept-fidelity. It \edit{further enables inference-time} combinations of concepts, and it has a small model size --- roughly 100KB per concept.
To achieve these goals, we focus on the cross-attention module of diffusion-based \TTI models.

In typical diffusion-based \TTI, an input text prompt is transformed into a sequence of encodings using a text encoder such as T5~\citep{raffel2020exploring} or CLIP~\citep{radford2021learning}. These encodings are then mapped to Keys and Values using learned projection matrices as part of the cross-attention module. Inspired by \citet{balaji2022ediffi, hertz2022prompt}, we propose to view the effects of these projections as two different pathways: The Keys (K) are a ``\Where'' pathway, which controls the layout of the attention maps, and through them the compositional structure of the generated image. The Values (V) are a ``\What'' pathway, which controls the visual appearance of the image components.

Our main insight is that existing techniques fail when they overfit the \Where pathway (Figures \ref{fig_attention_viz}, \ref{fig_lock_curves}), causing the attention on the novel words to leak beyond the visual scope of the concept itself.
To address this shortcoming, we propose a novel ``\textit{Key-Locking}'' mechanism, where the keys of a concept are fixated on the keys of the concept's super-category. For example, the keys that represent a specific teddybear may be key-locked to the super-category of a teddy instead. Intuitively, this allows the new concept to inherit the super-category's qualities and creative power (\figref{fig1}). Personalization is then handled through the \What (V) pathway, where we treat the Value projections as an extended latent-space and concurrently optimize them along with the input word embedding.

Finally, we describe how these components can be incorporated directly into the \TTI model through the use of a \textit{gated} rank-1 update to the weights of the K and V projection matrices. The gated aspect of this update allows us to combine multiple concepts at inference time by selectively applying the rank-1 update to only the necessary encodings. Moreover, the same gating mechanism provides a means for regulating the strength of learned concept, as expressed in the output images.
This allows runtime efficient, inference-time trade-off of visual-fidelity with textual-alignment, without requiring specialized models for every new operating point.
Empirically, \ourmethod not only leads to more accurate personalization at a fraction of the model size, but it also enables the use of more complex prompts and the combination of \textit{individually-learned} concepts at inference time.

In summary, our contributions are as follows: First, we investigate the overfitting observed in current personalization methods and propose a ``Key-Locking'' mechanism that mitigates it. Second, we propose a controllable rank-1 update mechanism for the network that achieves high object fidelity with only a 100KB footprint.
Third, our approach efficiently spans the Pareto front of a single trained model to balance visual-fidelity and textual-alignment during runtime, while also being able to generalize to unseen operating points.
Finally, we demonstrate that \ourmethod can outperform the state-of-the-art and enable object compositions at inference time.

\section{Related work}
\textbf{Diffusion based text-guided synthesis.}
Recent advances in \TTI generation have been led by pre-trained diffusion models~\citep{ho2020denoising} and particularly by large models~\citep{balaji2022ediffi,rombach2021highresolution,ramesh2022hierarchical,nichol2021glide} trained on web-scale data~\citep{schuhmann2021laion}.
Our approach extends these pre-trained models to portray personalized concepts.
It is applied with Stable-Diffusion \citep{rombach2021highresolution}, but we expect that it can be applied to any \TTI generator that uses a similar cross-attention mechanism~\citep{saharia2022photorealistic}.

\textbf{\TTI Personalization} The task of \TTI personalization~\citep{gal2022textual,ruiz2022dreambooth} aims to teach a generative \TTI model to synthesize new images of a specific target concept, guided by free language. Current personalization methods either fine-tune the denoising network around a fixed embedding~\citep{ruiz2022dreambooth} or optimize a set of word embeddings to depict the concept~\citep{gal2022textual,cohen2022my,agrawal21KnownUnknowns,Daras2022MultiresolutionTI}. 
\cite{kumari2022customdiffusion} is a concurrent work that fine-tunes the K and V cross-attention layers of the denoising network, along with a word embedding. It uses a closed-form optimization technique to combine concepts.
In \ourmethod, we lock the K pathway to the concept's supercategory and use gated rank-1 editing instead of fine-tuning and subsequent optimization. This yields novel quality of results. We focused on K and V pathways independently of \cite{kumari2022customdiffusion}.

\textbf{Rank-1 Model editing} In the field of natural language processing, significant effort was given to understanding and localizing the memory mechanisms of large language models. Specifically, it has been observed that the transformer feed-forward layers serve as key-value memory storage~\citep{geva2020transformer,geva2022transformer,meng2022locating}.
Recently, \citet{meng2022locating,bau2020rewriting} introduced ROME, a rank-1 editing approach that updates these associative memory layers in order to modify the network's factual knowledge.
\ourmethod seeks to apply similar ideas to \TTI diffusion models. However, \naive rank-1 editing of diffusion models can lead to poor results, as also reported by~\citep{kumari2022customdiffusion}. Our approach addresses the challenges of applying these methods to diffusion-based cross-attention layers. Moreover, \ourmethod can combine multiple rank-1 edits using a dynamic gating mechanism rather than a static edit.

\textbf{Text-based image-editing.} The advent of powerful multi-modal models has brought with it an array of text-based editing methods~\citep{bau2021paint,gal2021stylegan,patashnik2021styleclip}. With diffusion models, these range from single-image editing approaches~\citep{kawar2022imagic,Wu2022UncoveringTD,valevski2022unitune,Zhang2022SINESI,InstructPix2Pix,mokady2022null,meng2021sdedit} to inpainting tasks~\citep{nichol2021glide,ramesh2022hierarchical,Yang2022PaintBE,avrahami2022blended}. Most relevant to our work are the paint-with-words (PWW) approach introduced by \citet{balaji2022ediffi}, and prompt-to-prompt (P2P)~\citep{hertz2022prompt}. PWW biases the attention map toward a predefined mask during inference time. P2P edits a \textit{given} generated image, by regenerating it with a new prompt while injecting the attention maps of the original image along the diffusion process. In contrast to these methods, we do not edit \textit{given} images but learn to represent a personalized concept that can be invoked in new prompts. Additionally, we do not override the attention map, but constrain the cross-attention Keys of the new concept. These are a contributing factor to the attention map, but they still allow for concept-specific modifications through the Query features.

\section{Preliminaries and Notations}
We begin with an overview of two mechanisms that our work leverages for personalization. The first is the cross-attention mechanism typically found in \TTI diffusion models \cite{rombach2021highresolution}. The second is a recent approach for rank-1 editing of large language models \cite{meng2022locating}.

\subsection{Cross-Attention in Text-to-Image models}
\label{sec_notation}
In current \TTI systems based on diffusion models \edit{(Fig. \ref{fig_architecture}.A)}, an input text prompt is first converted into a sequence of word-embeddings. This sequence is then transformed into a sequence of encodings using a text encoder, such as CLIP \cite{radford2021learning}. Each encoding is then linearly projected through two cross-attention matrices: $\mW_K$ and $\mW_V$. The results of these projections are known as ``Keys" and ``Values".
Formally, let $M$ be the length of the input sequence, $\vw \in \R^{M \times d_w}$ be a sequence of word-embeddings, each with dimension $d_w$, and $\e \in \R^{M \times d_e}$ be a sequence of encodings, each with dimension $d_e$. For each entry $m \in M$ in the sequence, the encoding $\Em \in \R^{d_e}$ is mapped by the two projection matrices into a ``Key" vector $K_{m}=\mW_K e_m \in \R^{d_k}$ and a ``Value" vector $V_m = \mW_V e_m$. Concurrently, local image features are projected through a third matrix, $\mW_Q$, generating a spatial map of ``Queries". These are in turn projected onto the keys, yielding a per-encoding attention map: $A_m = \textit{softmax}\left({Q K_m^T}/{\sqrt{d_k}}\right)$~\cite{rombach2021highresolution}. Intuitively, this map informs the model about the relevance of the $m^{th}$ word to each spatial region of the image.
Finally, local image features are comprised by the ``Values'', weighted by these maps: $A \cdot V$ (Figure \ref{fig_architecture}).

\subsection{Rank-1 Model Editing}
\label{sec_ROME}

Rank-one Model Editing (ROME) \cite{meng2022locating} is a recently introduced method for editing factual association in a pre-trained language model, such as GPT. %
ROME edits the weights of a single linear layer $W$ in the network, so that given one \textit{target-input} $\istar$, the layer will emit one  \textit{target-output} $\ostar$ \footnote{Intuitively, the \textit{target-input} plays a role of a key that is to be matched. To avoid confusion with transformers' $K$ and $V$ pathways, we use ``target-input" and ``target-output" instead of ``key" and ``value" from \cite{meng2022locating}.}. %
To edit the model's factual knowledge, ROME employs three steps, performed separately.
(1) \textit{Find the target-input $\istar$ associated with the edited word (fact) in layer $W$.}
ROME determines the target-input activations of the edited word in different prompts by passing them through the language model and averaging the activations at the word's index. This gives the representation $\istar$.
(2) \textit{Find the target-output $\ostar$}, by optimizing the output activation of the layer for a specific goal --- e.g. to modify the facts presented by the model's final output. $\ostar$ is optimized over the output of a \textit{single} word index.%
(3) \textit{Update the layer $W$} by solving a constrained least-squares problem, which has a closed-form solution
\begin{align}
    \hat{W} = W + \Lambda (C^{-1}\istar)^T.
    \label{eq_ROME}
\end{align}
Here, $\Lambda = (\ostar - W \istar)/[(\istar^T (C^{-1})^T \istar)]$, $C$ is a constant positive definite matrix, that is pre-cached (Appendix \ref{sec_supp_additional_details}). The update in step 3 is performed after finding $\ostar$, thus affecting all sequence encodings, instead of just the single edited word as in step 2.

By limiting the update to a local, rank-1 change, ROME changes the information associated with a single fact without drastically altering the knowledge of the model.
Our method leverages a similar mechanism to edit a text-to-image model and introduce new visual concepts.

\section{Method} \label{sec_method}

\begin{figure}[h]
\includegraphics[width=\columnwidth, trim={1.6cm 23.1cm 12cm 4.2cm},clip]{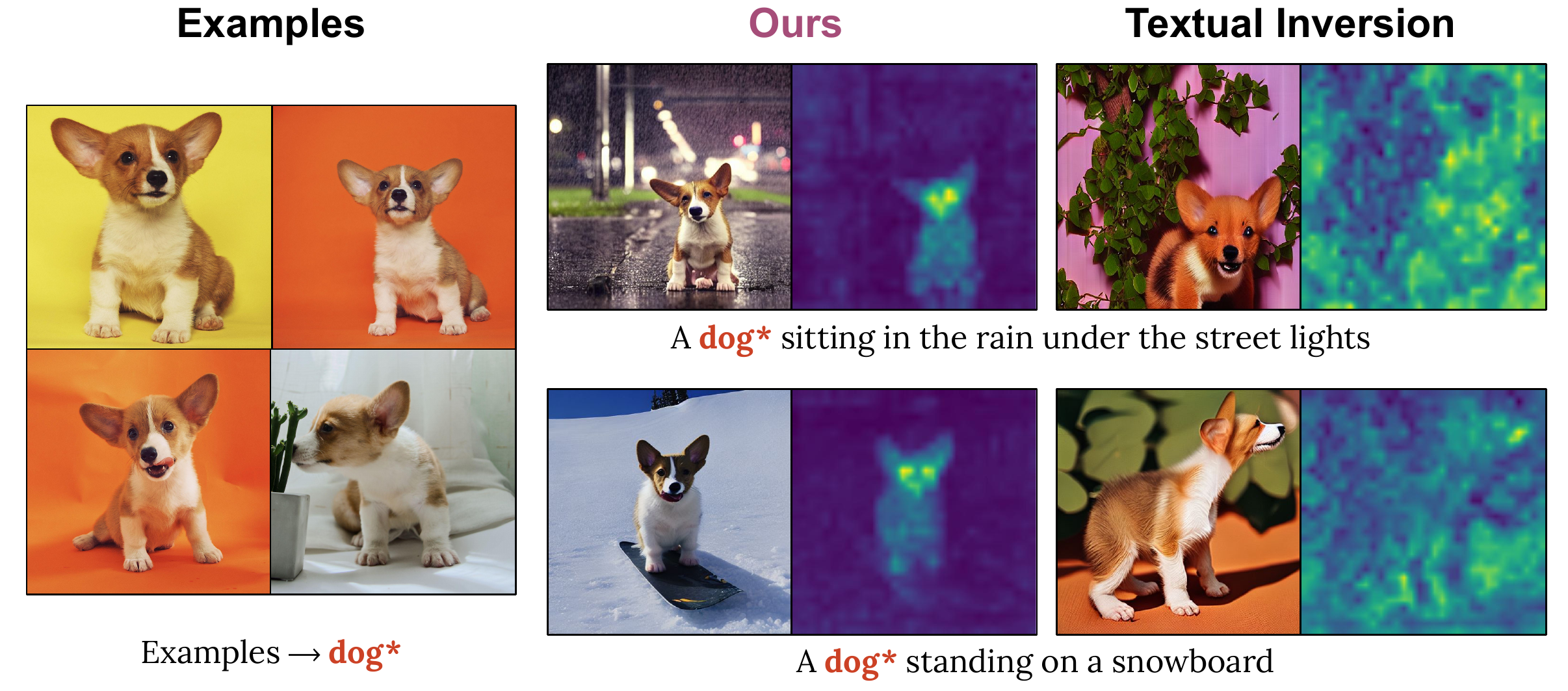} %
    \caption{ \textbf{Attention overfit:}
    Typical overfit in Textual-Inversion (TI), caused by the attention of the learned embedding taking over the whole image. Here we visualize the attention maps that correspond to the ``dog*" word. The TI attention regions (right panel) are spread across the entire image rather than focusing on the object. This leads the generative process to ignore the rest of the prompt and depict only the ``dog*" concept.
}
\label{fig_attention_viz}
\vspace{-5pt}
\end{figure}

We aim to personalize a model in an expressive and efficient manner. A natural place to start then is by investigating the limitations of prior work in the field, and particularly \textit{Textual Inversions}~\citep{gal2022textual} and \textit{DreamBooth}~\citep{ruiz2022dreambooth}. We notice that these methods, and Textual Inversion in particular, are susceptible to overfitting, where a learned concept becomes difficult to modify by changing the prompt that contains it. In \figref{fig_attention_viz} we demonstrate that this issue originates in the attention mechanism, \edit{as the new concept draws attention beyond its visual scope. Additional examples showing this phenomenon are provided in \figref{fig_attention_appendix}.}

Next, we describe \textbf{\textit{\ourmethod}}, an approach to overcome the problem through rank-1 layer editing. We outline a gating mechanism that provides better control at inference time and describe how to leverage it to compose concepts that were learned in isolation.

\begin{figure*}[t]
\centering
\includegraphics[width=0.97\textwidth, trim={0.cm 0.1cm 0.7cm 0.cm},clip]{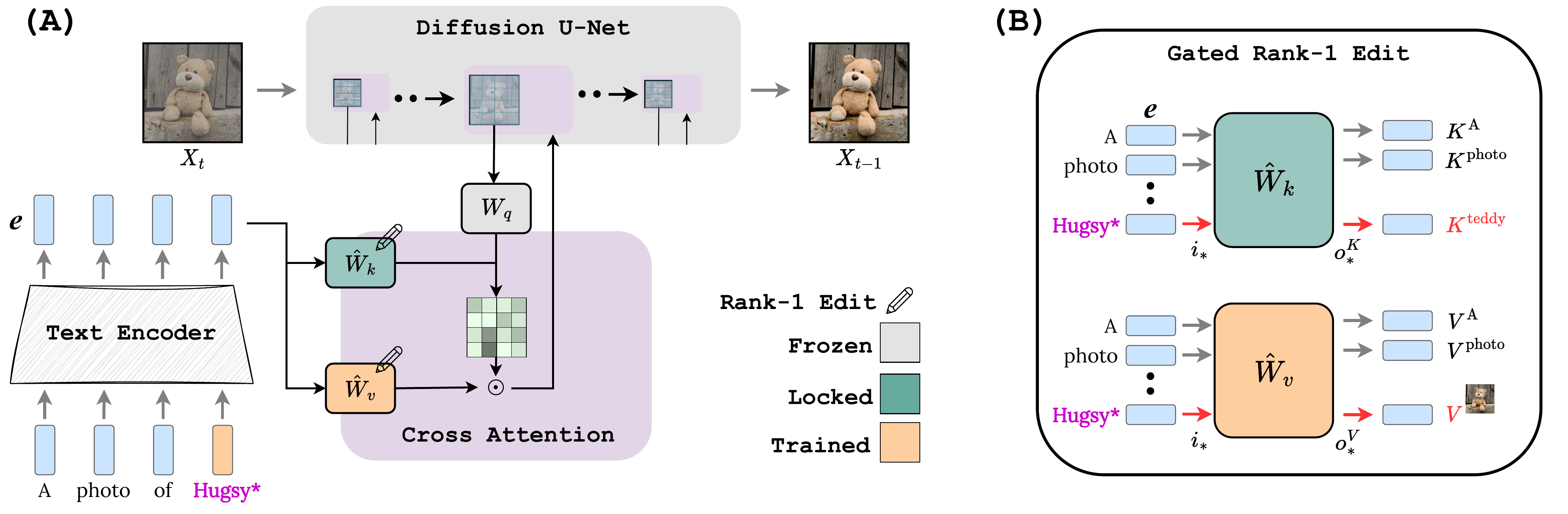} %
    \caption{
    \textbf{Architecture outline (A):} A prompt is transformed into a sequence of encodings. Each encoding is fed to a set of cross-attention modules (purple blocks) of a diffusion U-Net denoiser. Zoomed-in purple module shows how the Key and Value pathways are conditioned on the text encoding. The Key drives the attention map, which then modulates the Value pathway. %
    \textbf{Gated Rank-1 Edit (B):}  \textit{Top:} The K pathway is locked so any encoding of $e_\text{Hugsy}$ that reaches $\hat{W}_k$ is mapped to the key of the super-category $K^\text{teddy}$. \textit{Bottom:} Any encoding of $e_\text{Hugsy}$ that reaches $\hat{W}_v$, is mapped to $V^\text{Hugsy}$, which is learned. The gated aspect of this update allows to selectively apply it to only the necessary encodings and provides means for regulating the strength of learned concept, as expressed in the output images.
    }
    \label{fig_architecture}
\vspace{-5pt}
\end{figure*}

\subsection{Two conflicting goals and one \Naive Solution}
Personalized \TTI aims to achieve two goals:
(1) \textit{Avoid overfitting} to the example images, so the personalized concept can be generated in various poses, appearances, or context; and (2) \textit{Preserve the identity} of the personalized concept in the generated image, despite being portrayed in a different pose appearance or context. There is a natural trade-off between these two goals. Methods that overfit the input examples tend to preserve identity, but then fail to match creative prompts that aim to place the concept in different contexts.

\textbf{The \Where Pathway and the \What Pathway}. To improve both of these goals simultaneously, our key insight is that models need to disentangle what is generated from where it is generated. To this end, we leverage the interpretation of the cross-attention mechanism described in section \ref{sec_notation}. The $K$ pathway --- the one associated with the ``Keys", is related to creating the attention map. It thus serves as a pathway for controlling \textit{where} objects are located in the final image. In contrast, the $V$ pathway is responsible for the features added to each region. In this sense, it can control \textit{what} appears in the final image.
We therefore interpret $K$ mappings as a ``\Where'' pathway and  $V$ mappings as a ``\What'' pathway, and this interpretation guides our proposed method:

\subsubsection{Avoid overfitting}
In preliminary experimentation, we noticed that when learning personalized concepts from a limited number of examples, the model weights of the \Where pathway ($\mW^K$) are prone to overfit to the image layout seen in these examples. \figref{fig_attention_viz} illustrates this problem showing that the personalized examples may `dominate' the entire attention map, and prevent other words from affecting the synthesized image. We thus aim to prevent this attention-based overfitting by restricting the \Where pathway.

\subsubsection{Preserving Identity.}
In Image2StyleGAN, \citet{abdal2019image2stylegan} proposed a hierarchical latent-representation to capture identities more effectively. There, instead of predicting a single latent code at the generator's input space, they predicted a different code for each resolution in the synthesis process. We propose the \What (V) pathway activations as a similar latent space, given their compact nature and the multi-resolution structure of the underlying U-Net denoiser.

\subsubsection{A \Naive Solution.}
To meet both goals, consider this simple solution: Whenever the encoding contains the target concept, ensure that its cross-attention \textit{keys} match those of its supercategory, which we call \textit{\textbf{Key Locking}}. Additionally, we want the cross-attention \textit{values} to represent the concept in the multi-resolution latent space. For example, as illustrated in \figref{fig_architecture} (right-top), given image examples of our teddy bear named Hugsy, when the encoding includes ``Hugsy" the V projection emits a concept-specific code, while the K projection is targeted to emit keys for the super category $K^\text{teddy}$.

One way to implement this idea would be a simple vector replacement - simply swapping out the keys and values assigned to the encoding at the personalized concept's index. However, this fails to account for the cross-word information sharing in the text encoder. By the time the encoding reaches the denoiser's cross-attention layers,  its features are already influenced by the features of other words in the text, and in turn, influence them as well. We want to ensure that our implementation accounts for this influence, and correctly modifies the Key and Value activations for any such influenced words.

A natural solution is then to edit the weights of the cross-attention layers, $\mW_V$ and $\mW_K$ using ROME. Specifically, when given a target-input $i_\text{Hugsy}$ we enforce the $K$ activation to emit a specific target-output $o^K_\text{Hugsy} = K^\text{teddy}$. Similarly, given a target-input $i_\text{Hugsy}$, we enforce the $V$ activation to emit a learned output $o^V_\text{Hugsy} = V^\text{Hugsy}$ see \figref{fig_architecture} (right-bottom). Now, for any word, if its encoding contains a component parallel (aligned) to $\istar$, then their activation outputs will also be modified accordingly.

Unfortunately, applying ROME to this task faces two challenges: \\
\textbf{Challenge 1:} Training with ROME leads to a mismatch between training and inference. This is because during training in ``step 2'', ROME optimizes only the target-output $\ostar$ associated with one specific \edit{entry in the prompt} ($m^{th}$-index). However, as noted above, when performing the rank-1 matrix update in ``step 3'' the change is expected to affect the projections of other words in the prompt. Indeed, we have observed that this results in a train-test mismatch that substantially degrades the fidelity of the reconstructed concept. \\
\textbf{Challenge 2:} A similar effect also prevents us from combining more than one learned concept, as their effects on the projections are not well-disentangled. Moreover, these new concepts are associated with multiple target-inputs $\istar$, which may themselves be inherently entangled (e.g. if the concepts share related semantics). Together, these lead to the creation of visual artifacts when attempting to combine concepts at inference-time.

To address these challenges we propose to align the training and inference steps of ROME, and introduce a new gating mechanism. Both components are described below.

\subsection{Gated Rank-1 Model Editing for Personalized \TTI}

\textbf{Training end-to-end to address train-test mismatch.} To address the first challenge, ROME's mismatch between training and inference, we propose to unify the second and third steps of ROME. As such, the target-output optimization and matrix update occur together during training. The network learns to account for any effects on other prompt-parts, avoiding the train-inference mismatch.

To do so, we rewrite the weight update of ROME, to characterize the output $h$ of layer $\hat{W}$ when presented with an input $\Em$. This yields
\begin{align}
        h = W \Em^\perp + {\ostar sim(\istar, \Em)}/{||\istar||^2_{C^{-1}}},
   \label{eq_ROME_sim}
\end{align}
Here, $sim(\istar, \Em) := \istar^T (C^{-1})^T \Em$ measures the similarity of \Em with $\istar$ in a metric space defined by $C^{-1}$ \cite{atzmon2015learning}, $||\istar||^2_{C^{-1}} := sim(\istar, \istar)$ measures the energy of $\istar$ in the same metric space, and $\Em^\perp := \Em - {\istar \textit{sim}(\istar, \Em) }/{||\istar||^2_{C^{-1}}}$ is the component of $\Em$ that is orthogonal to $\istar$ in the metric space. \edit{Intuitively, the right additive term in \eqref{eq_ROME_sim} maps the $\istar$ component of the word encoding ($e_m$) to $\ostar$. The left term nulls the $\istar$ component from the word encoding and maps the remaining using the pretrained matrix $W$.}  We provide more details in Appendix \ref{sec_supp_lemma} and \ref{supp_single_orthogonal}.

Given this characterization, we replace the forward pass of each layer by updating it using \eqref{eq_ROME_sim} as the layer's forward pass.
This ensures that the same update expression is used for both training and inference, eliminating the mismatch. Combined with an online estimation of $\istar$ (\secref{sec_implementation_details}), it enables end-to-end training with ROME, rather than individually applying its 3 steps.

\textbf{Using gated rank-1 update for combining concepts.}
Combining individually learned concepts at inference time is a hard challenge.
In initial experiments, we tested adding concepts one by one,
by editing $W$ using $\hat{W}= W+\sum_{j=1..J} \Lambda^j (C^{-1}\istar^j)^T$ as a variation of equation \eqref{eq_ROME}.
We found this approach introduced visual artifacts, even when the prompt only included a single learned concept. We hypothesize that this problem arises because the different $\istar$s of individually learned concepts may interfere with each other. For example, they may have related representations if the concepts share semantics.
Ideally, \edit{a concept update occur only when input encoding has sufficient energy regarding the concept}, and attenuated otherwise. By doing so, we can ensure that the model update is only applied to the relevant concept, and not to others.

To address this challenge, we use a \textit{gating} mechanism to selectively allow or attenuate the influence of each concept on the layer output. Here, we note that the update rule of \eqref{eq_ROME_sim} already includes a \textit{linear} gating mechanism $\textit{sim}(\Em, \istar)/{||\istar||^2_{C^{-1}}}$, which is close to $1$ when $\Em = \istar$. However, lower similarity values may not be sufficiently attenuated.
We therefore propose to increase the influence of $\textit{sim}$ by wraping the $\textit{sim}(\Em, \istar)/{||\istar||^2_{C^{-1}}}$ value with a sigmoid function, which has hyper-parameters for bias and temperature. This way, the weight updates are sharply concentrated on inputs that strongly correspond to the personalized concept.

Therefore, the forward pass of each layer update during both training and inference time of a single concept, is
\begin{align}
     h = W \Em^\perp + {\ostar \sigma\left(\frac{\textit{sim}(\istar, \Em)/{||\istar||^2_{C^{-1}}} - \beta}{\tau} \right) }
    \label{eq_gated_single}
\end{align}
where, $\sigma$ is a sigmoid activation function, $\tau$ is the temperature and $\beta$ a bias term. This implementation ensures that weight updates are only applied to encoding components that align (parallel) with $\istar$, \ie those belonging to the new concept, or influenced by it.

This non-linear gating mechanism therefore provides \textbf{two important benefits}: First, it allows us to better separate the influence of individually learned concepts during inference time. Second, even for a single concept, it allows for \textit{inference-time} control over the influence of the concept. By adjusting the values of the sigmoid hyper-parameters, the bias and the temperature, we can trade visual fidelity with textual alignment and vice-versa.

In the next section we expand on how to generalize this formulation to combine multiple concepts.

\subsection{Inference}
\textbf{Single concept:}
For inference with a single trained concept, we simply apply \eqref{eq_gated_single} to the forward pass of each edited cross-attention layer. We can control the strength of the depicted concept by changing the values of the sigmoid's $\tau$ and $\beta$ at inference time.

\textbf{Combining multiple concepts:}
To combine concepts that were trained in isolation, we extend equation \eqref{eq_gated_single} to include multiple concepts $\{\istar^j, \ostar^j \}_{j \in 1 ..J}$. For that, we first generalize $\Em^\perp$ to be orthogonal to the sub-space spanned by all the $\{\istar^j\}_{j=1..J}$ in the metric space, which we denote as $\Em^{\perp J}$. For the right term we simply sum the gated responses from all the concepts. The final expression is:
\begin{align}
     h = W \Em^{\perp J} + \sum_{j \in 1..J}{\ostar^j \sigma\left(\frac{\textit{sim}(\istar^j, \Em)/{||\istar^j||^2} - \beta_j} {\tau} \right)},
    \label{eq_gated_multi}
\end{align}
where $\Em^{\perp J} = \Em - \sum_{j \in 1..J}{u_j\textit{sim}(u_j, \Em) }$, and $u_j$ relates to a basis vector in the metric space, after being projected back to the text encoder space by an inverse Cholesky root ${L^T}^{-1}$. The derivation of $\Em^{\perp J}$ is provided in Appendix \ref{supp_span_orthogonal}.

\subsubsection{Global Key-Locking:}
Key-Locking ensures that a concept's Key is correctly aligned with its superconcept. However, it does not ensure that the text-encoder handles the concept in the same way it would have handled the superconcept and its correlations to the other words in the encoding. We also investigate an inference time method to align Key-locked concepts to an entire prompt. We refer to this variant as \textit{global} key-locking, and to our vanilla mechanism as \textit{local} key locking. We describe the details in Appendix \ref{sec_supp_additional_details}.

\begin{figure*}[!ht]
    \centering
    \includegraphics[width=0.88\textwidth, trim={2.0cm 8.8cm 12.2cm 2.3cm},clip]{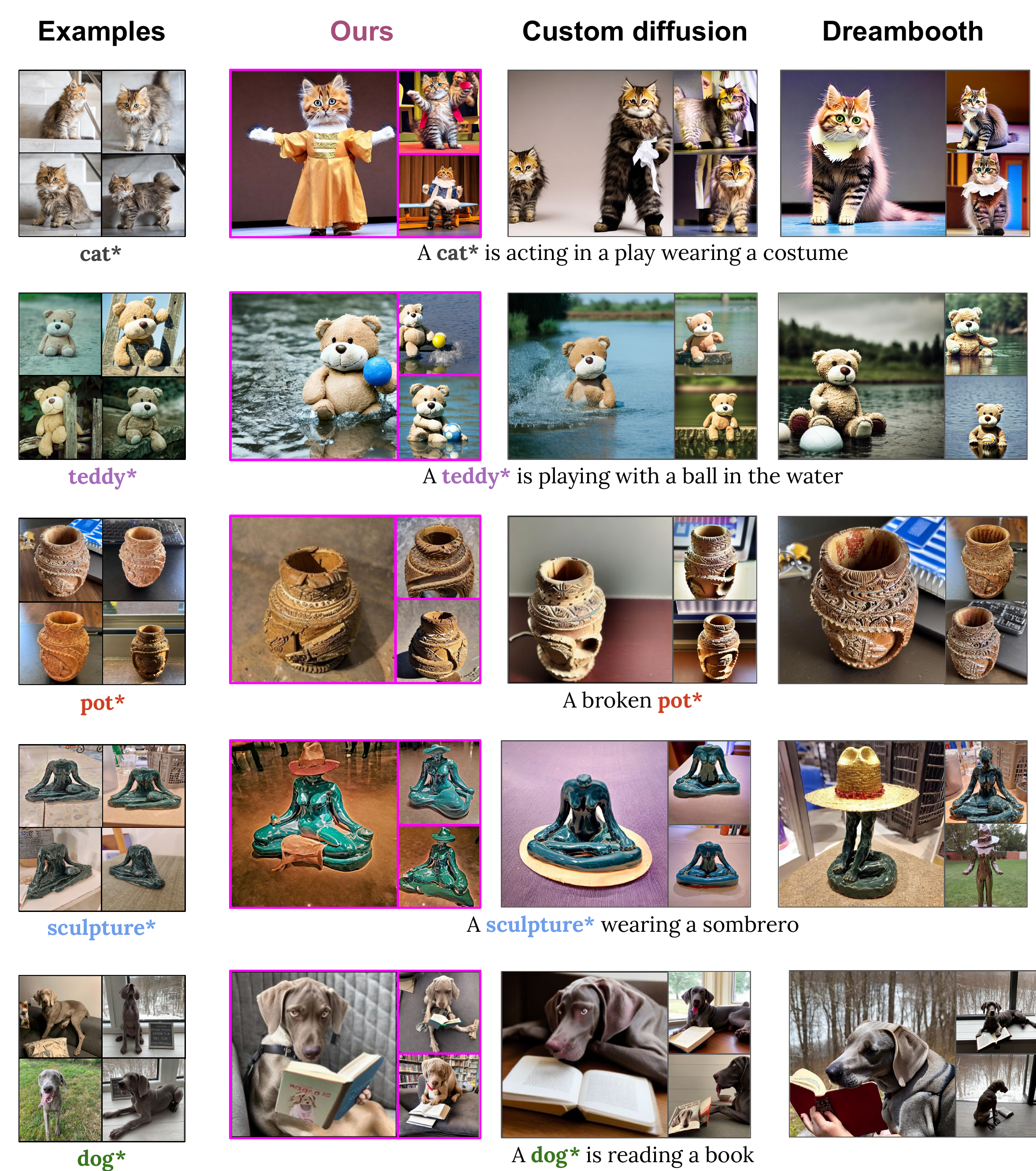} %
    \caption{
    \textbf{Generation results with single concept examples.}
    For each concept, we show exemplars from our training set, along with generated images, their conditioning texts and comparisons to Custom-Diffusion (CD) and Dreambooth (DB) baselines. \ourmethod can enable more animate results, with better prompt-matching and less susceptibility to background traits from the original image. Note in particular the improved garments and theatrics on our cat (top), or the prompt-appropriate gaze and posture when instructing our dog to read a book (bottom). For some prompts, the baselines simply copy the content from the training set (\eg the pot).
    }
    \label{fig_single}
\vspace{-10pt}
\end{figure*}

\begin{figure*}[t]
    \includegraphics[width=\textwidth, trim={0.7cm 23.5cm 2.8cm 1.3cm},clip]{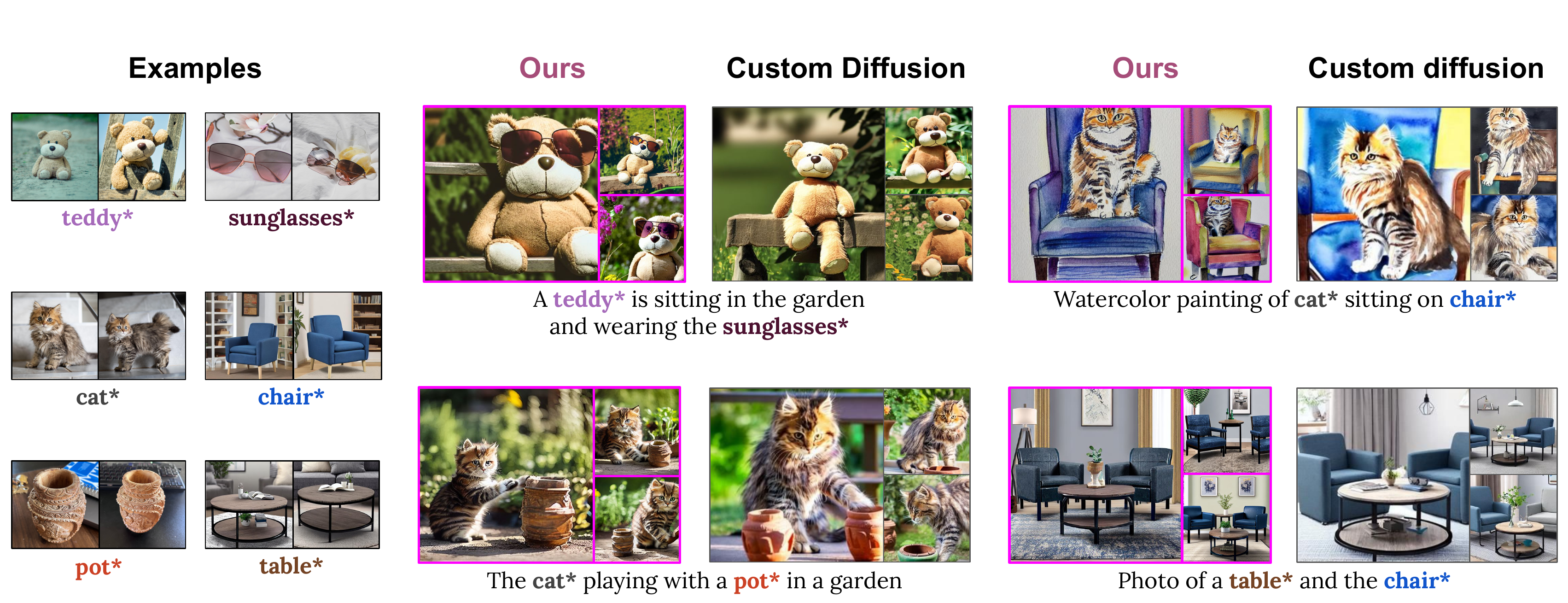} %

    \caption{
    \textbf{Additional generation results with multi concept examples.}
    We show pairs of concepts interacting, and compare to CD. Except for the teddy* prompt, all prompts are from CD paper and use the images provided by the paper. In the teddy* example, \ourmethod portrays it with the sunglasses*, while CD omits the sunglasses*. In the watercolor painting \ourmethod better preserve the chair shape. In the table* example, \ourmethod better preserve the table color.
    }
    \label{fig_multi}

\end{figure*}

\subsection{Implementation details}
\label{sec_implementation_details}

\myparagraph{Online estimation of $\istar$:}
We use the following exponential moving average expression to estimate $\istar$ during training time: $\istar := 0.99\istar + 0.01\ensuremath{\e_\text{concept}}$ where $\ensuremath{\e_\text{concept}}$ corresponds to encoding of the concept word at the output of the text encoder.

\myparagraph{Pseudo Code:} Appendix \ref{supp_pseudo_code} provides pseudo code for the rank-1 editing module of \ourmethod.

\myparagraph{Zero-Shot Weighting Loss:}
Training with few image examples is prone to learning spurious correlations from the image background. To decorrelate the concept from its background we weigh the standard conditional diffusion loss by a soft segmentation mask attained from a zero-shot image segmentation model \cite{luddecke2021prompt}. Mask values are normalized by their maximum value.

\myparagraph{Applying \ourmethod to multiple layers:} Similar to \cite{gal2022textual}, for each concept we choose a single word for a supercategory name. We use that word to initialize its word embeddings and treat the embeddings as learned parameters. We apply \ourmethod editing to all cross-attention layers of the UNet denoiser. For each of the K pathway layers ($l$), we precompute and freeze the $\ostar^{K:l}$ to be $\ostar^{K:l} = \mW^l_K \ve_\text{superclass}$  with a prompt saying ``A photo of a <superclass\_word>'', and we update $\istar$ as training progresses. On each of the V pathway layers, we treat $\ostar^{V:l}$ as learned parameters. %

\myparagraph{Training details:} We train $\ostar$ with a learning rate of $0.03$, for the embedding we set a learning rate of $0.006$. We use a batch size of $16$ using Flash-Attention \cite{dao2022flashattention,xFormers2022}. We only use flip augmentations $p=50\%$. We do not flip asymmetric objects. We use a validation set of 8 prompts, sampled every 25 training steps, and select the step with the model that maximizes the harmonic mean between a CLIP image similarity score, and a CLIP text similarity score. We describe the CLIP metrics in more detail in the experimental details. To condition the generation, we randomly sample neutral context texts, derived from the CLIP ImageNet templates~\citep{radford2021learning}. The list of templates is provided in the supplementary materials.

Our approach is trained on a single A100 GPU for an average of 210 steps, taking $\sim \!\!4$ minutes, and a maximum of 400 steps ($\sim \!\!7$ minutes). This training requires $\times2- \times3$ less compute compared to concurrent work \cite{kumari2022customdiffusion} and utilizes 27GB of RAM.

\myparagraph{Sigmoid hyper-parameters:}
\edit{At training time}, we set the sigmoid bias and temperature to $b=0.75, T=0.1$. At inference time we \edit{typically} use a temperature of $0.15$ and bias of $0.6-0.75$ for local key lock, or $0.4-0.6$ for global key lock.

\begin{figure}[h]
    \centering
    \includegraphics[width=0.9\columnwidth, trim={0.cm 0cm .4cm .5cm},clip]{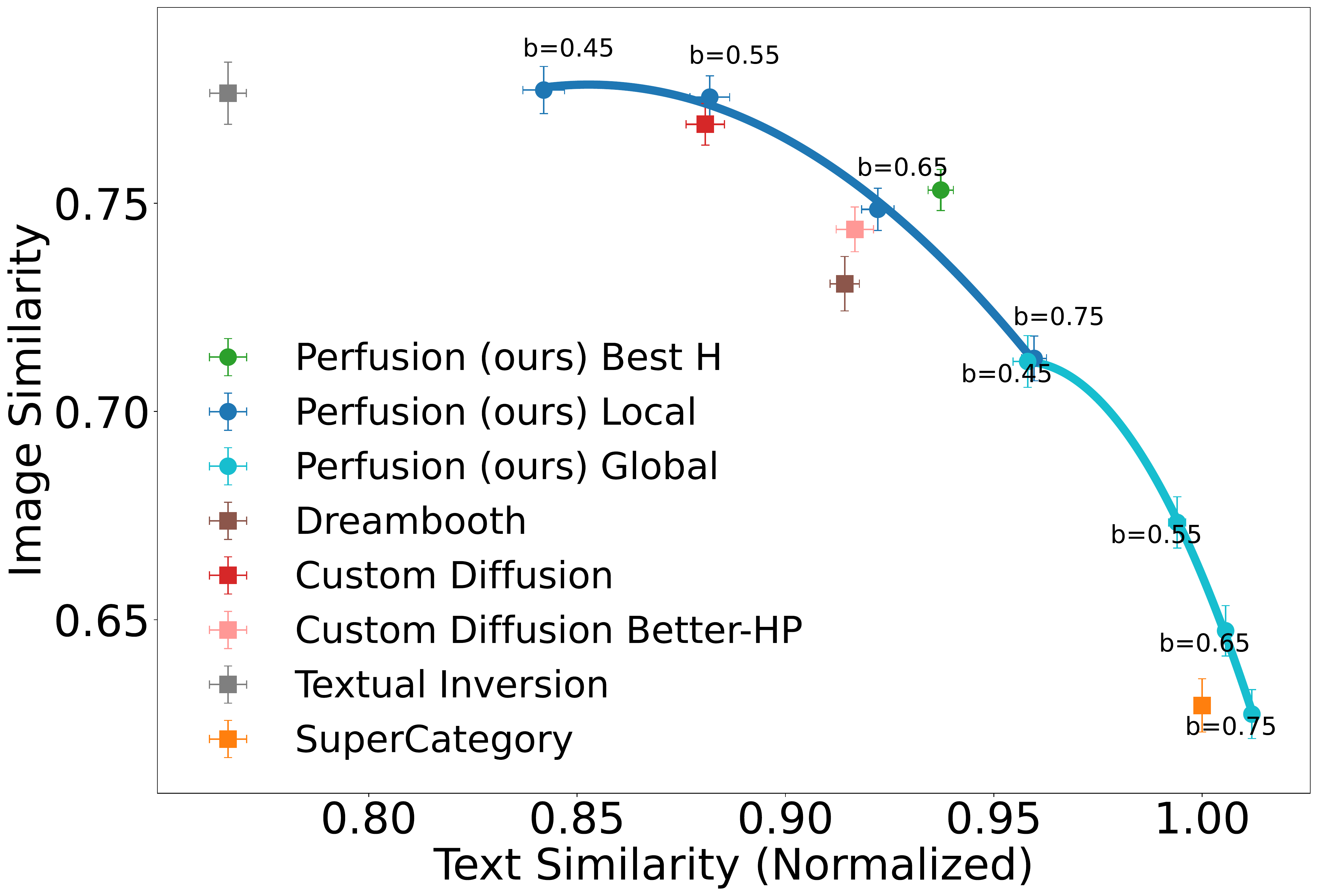} %
    \caption{ \textbf{Visual - Textual Similarity Plane:} With just a single 100KB trained model and run-time parameter choices, \ourmethod (blue and cyan) spans the Pareto front. Error bars denote 95\% confidence intervals.}
    \label{fig_main_pareto}
    \vspace{-5pt}
\end{figure}

\begin{figure}[t]
\vspace{-0pt}
\includegraphics[width=\columnwidth, trim={-0.2cm 26.2cm 15.7cm 0.3cm},clip]{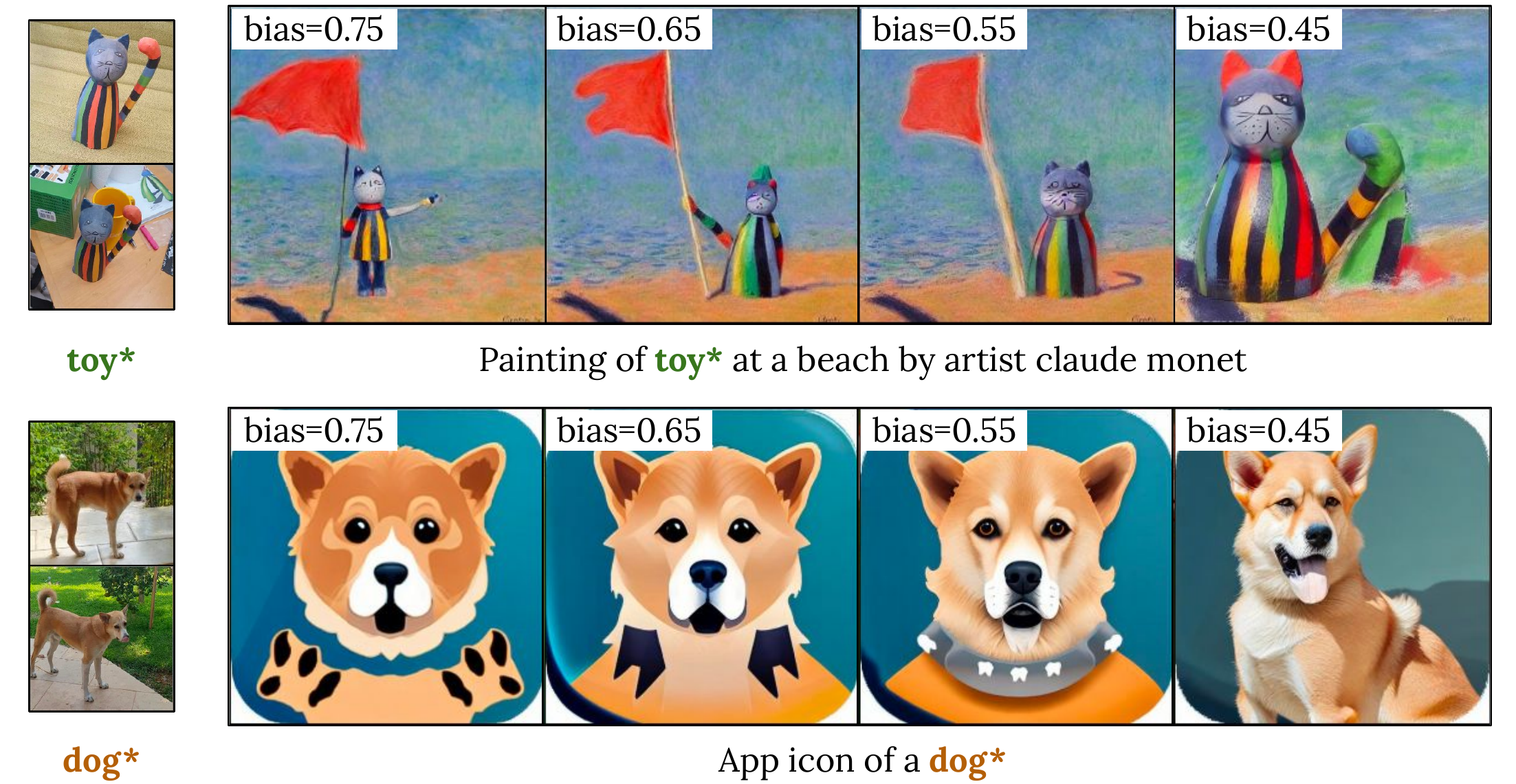} %
    \caption{ \textbf{Balancing visual-fidelity and textual-alignment:} Controlling the bias threshold of the sigmoid allows to balance the trade-off between visual-fidelity and the textual-alignment. A high bias value reduces the concept's effect, while a low bias value makes it more influential.
}
\label{fig_bias}
\vspace{-5pt}
\end{figure}

\begin{figure}[h]
\vspace{-0pt}
\includegraphics[width=\columnwidth, trim={1.4cm 2cm 0.6cm 1cm},clip]{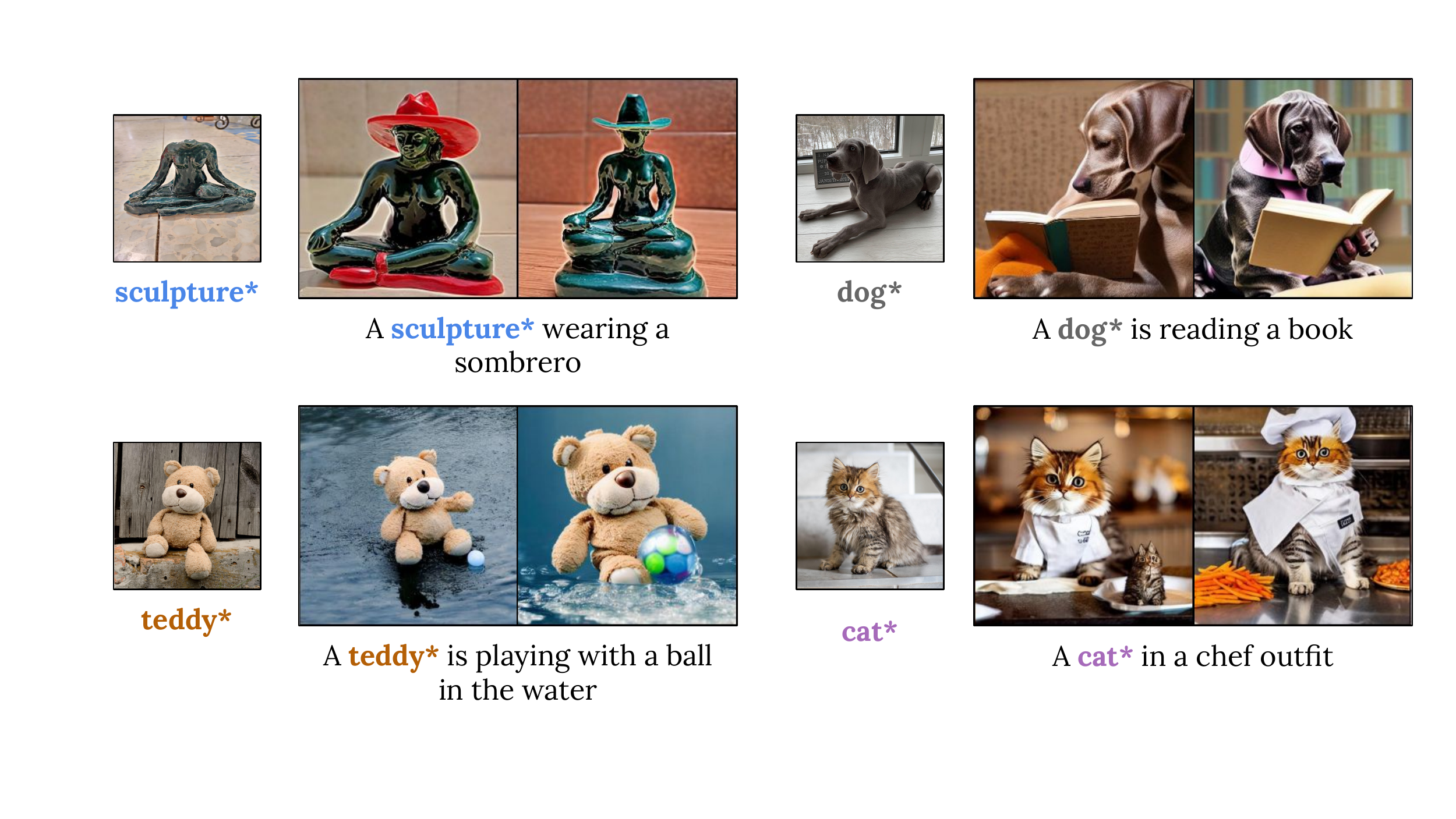} %
    \caption{ \textbf{1-shot training results:} When training with a single image, our method can generate images with both high visual-fidelity and textual-alignment. We provide quantitative results in \figref{fig_ablation_all}(a).  
}
\label{fig_oneshot_qualitative}
\vspace{-5pt}
\end{figure}

\section{Experiments}

We demonstrate that \ourmethod outperforms strong baselines. %
We conduct both a qualitative comparison and a quantitative evaluation, demonstrating that it spans the visual-fidelity and text-alignment Pareto front and achieves higher fidelity results with more complex prompts, despite using only a fraction of the parameters. Then, we study the properties of \ourmethod through an ablation study (\secref{sec_supp_ablation}).

\myparagraph{Compared Methods:}  %
\textbf{(1) \ourmethod}: Our method as described in \secref{sec_method}. We use a \textit{single} trained model for each concept, but we show results spanning different sigmoid biases and with \textit{local} and \textit{global} locking. These parameter adjustments are all applied during runtime. \textbf{(2) \ourmethod Best H}: For each class, we automatically choose the run-time variant with the best harmonic mean over text and visual similarities (see metrics).
\textbf{(3) DreamBooth} \cite{ruiz2022dreambooth}: A SoTA approach that fine-tunes all parameters of the denoiser's U-Net. We use the implementation of \citet{huggingface2022dreambooth}. \textbf{(4) Textual-Inversion (TI)} \cite{gal2022textual}: A SoTA approach that only optimizes word embeddings. We use the Stable Diffusion implementation from the official repository, with the parameters the authors report for LDM~\citep{rombach2021highresolution}.
\textbf{(5) Custom-Diffusion (CD)}: A concurrent work that trains the K and V cross-attention pathways, and also the word-embedding. We use the official implementation and hyperparameters. \textbf{(6) Custom-Diffusion Better-HP:} CD trained with 200 steps, which we found to improve the text similarity score. \textbf{(7) SuperCategory:} A text only baseline; We replace the concept word with its super class.
All methods were applied to a pre-trained Stable Diffusion checkpoint $v1.5$~\cite{rombach2021highresolution}.

\myparagraph{Dataset:}
\textit{Concepts:} For fair and unbiased evaluation we used concepts from previous papers: $6$ concepts from CD, $2$ from TI, and $3$ from or similar to DB, for a total of $11$ personalization concepts. These are from four groups: $4$ toys/figurines and $3$ pets (grouped as ``animated''), $2$ containers, $1$ furniture, and $1$ wearable accessory.
\textit{Prompts:} We use two types of prompts. First, $19$ prompts that were shared across all concepts. These only change the scene, but do not deform the concept. \edit{We name them ``\textit{shape-preserving}'' prompts}. Second, \textit{per-group} prompts. These are more specific to the group the concept belongs, and often induce a deformation to the concept appearance, like ``A \textit{broken} pot*", or ``A cat* is \textit{acting} in a play \textit{wearing a costume}". In total, we have $86$ unique prompts, with an average of $43$ prompts per class. Out of the $86$ unique prompts, $42$ were randomly selected from those used by Custom-Diffusion.

\myparagraph{Evaluation Metrics:}
Like TI, we report the results on a 2D plane to illustrate the balance between visual and text similarity. \edit{But unlike TI, which only uses prompts like "a photo of a $S^*$" to evaluate image similarity, we use all shape-preserving prompts to also measure concept fidelity in new scenes.}.

We compute the following evaluation metrics: \textbf{(1) Image similarity} is the average pairwise CLIP cosine-similarity between the concept images and the generated images from the shape-preserving prompts. \textbf{(2) Text Similarity} is the average CLIP similarity between all generated images and their textual prompts, where we omit the placeholder $S^*$ (i.e. "A
is dressed like a wizard"). To calibrate between different prompts,
we normalize text scores by comparing them to scores of images generated with a supercategory word instead of the learned word. \edit{All similarity scores are balanced ``per-class'' \cite{DRAGON}. Namely, we compute the mean score per concept class, and then average all class scores}. For each concept, we sample $8$ images per prompt, by $50$ DDIM steps and a guidance scale of $6$.

\subsection{Results}
In \figref{fig_main_pareto}, we illustrate the results on a plane that shows the trade-off between visual and textual similarity.
\textit{Dreambooth} successfully balances compromise both visual and textual fidelity. However, it obtains lower scores on both metrics when compared to \ourmethod and CD. \textit{Textual-Inversion} struggles to generalize to new prompts, showing low textual-similarity score. \textit{Custom-Diffusion} tends to favor visual similarity, even at the cost of overfitting to the target. We observe that with a small hyper-parameter change, \textit{Custom-Diffusion Better-HP}, can instead be balanced toward textual similarity.
\ourmethod outperform these baselines and pushes forward the pareto front. Notably, we can span this front using inference-time parameter modifications, which allows a user to control this trade-off based on their desired qualities.
In practice, this allows us to easily select the best operating point for each class, leading even greater performance (\textit{\ourmethod Best H}). Note that \ourmethod achieves these gains while requiring only 100KB of per-concept parameters, compared to several GBs in DB and nearly 100MB in CD.
\textbf{Importantly}, the comparison reveals that Key-Locking significantly improves textual similarity without significant harm to the visual similarity.

\begin{figure}[t]
\begin{overpic}[width=0.49\columnwidth, trim={0.cm 0cm .32cm .3cm},clip]{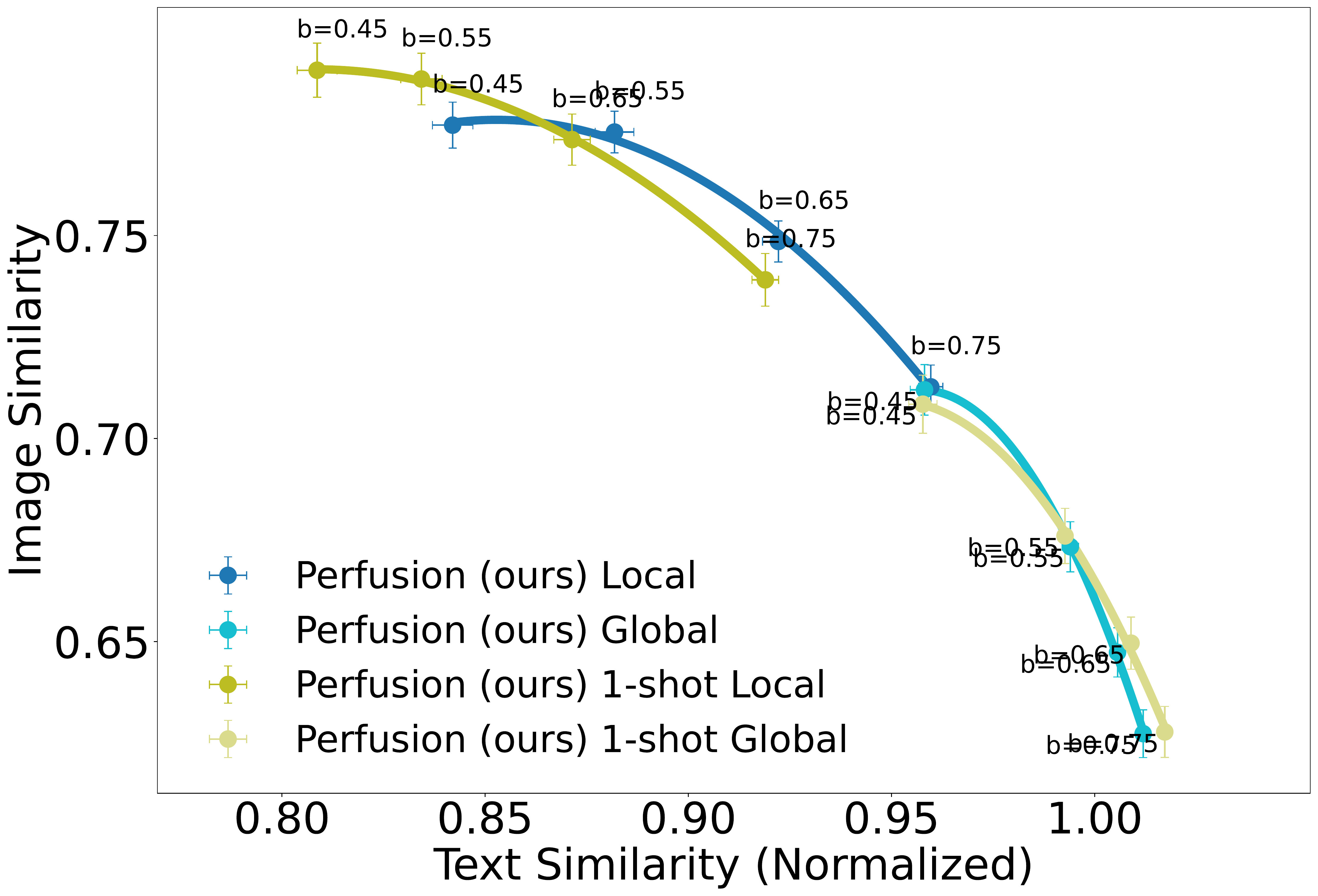} %
    \put (90,60) {(a)}
    \end{overpic}
\begin{overpic}[width=0.49\columnwidth, trim={0.cm 0cm .32cm .3cm},clip]{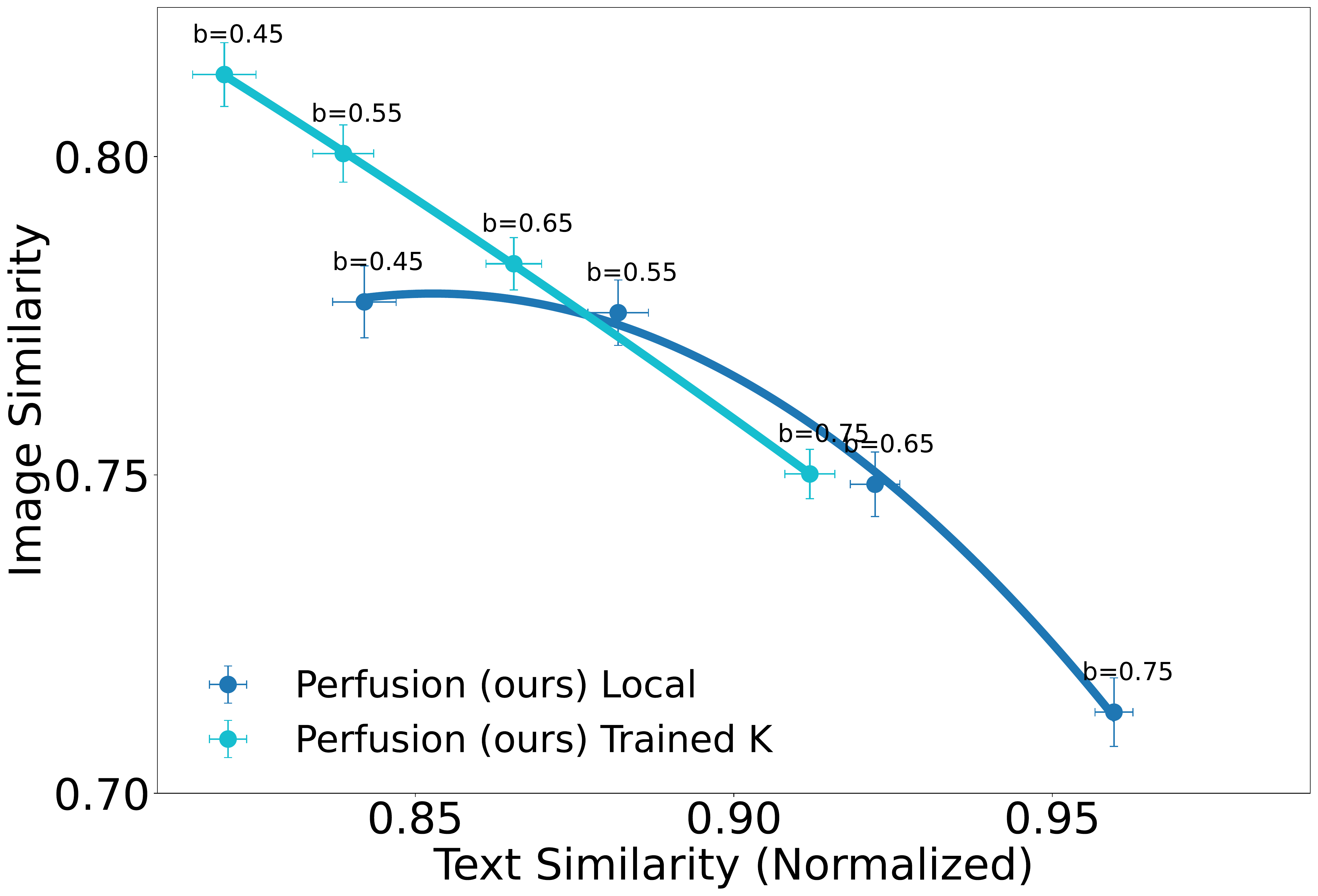} %
    \put (90,60) {(b)}
    \end{overpic}

\begin{overpic}[width=0.49\columnwidth, trim={0.cm 0cm .32cm .3cm},clip]{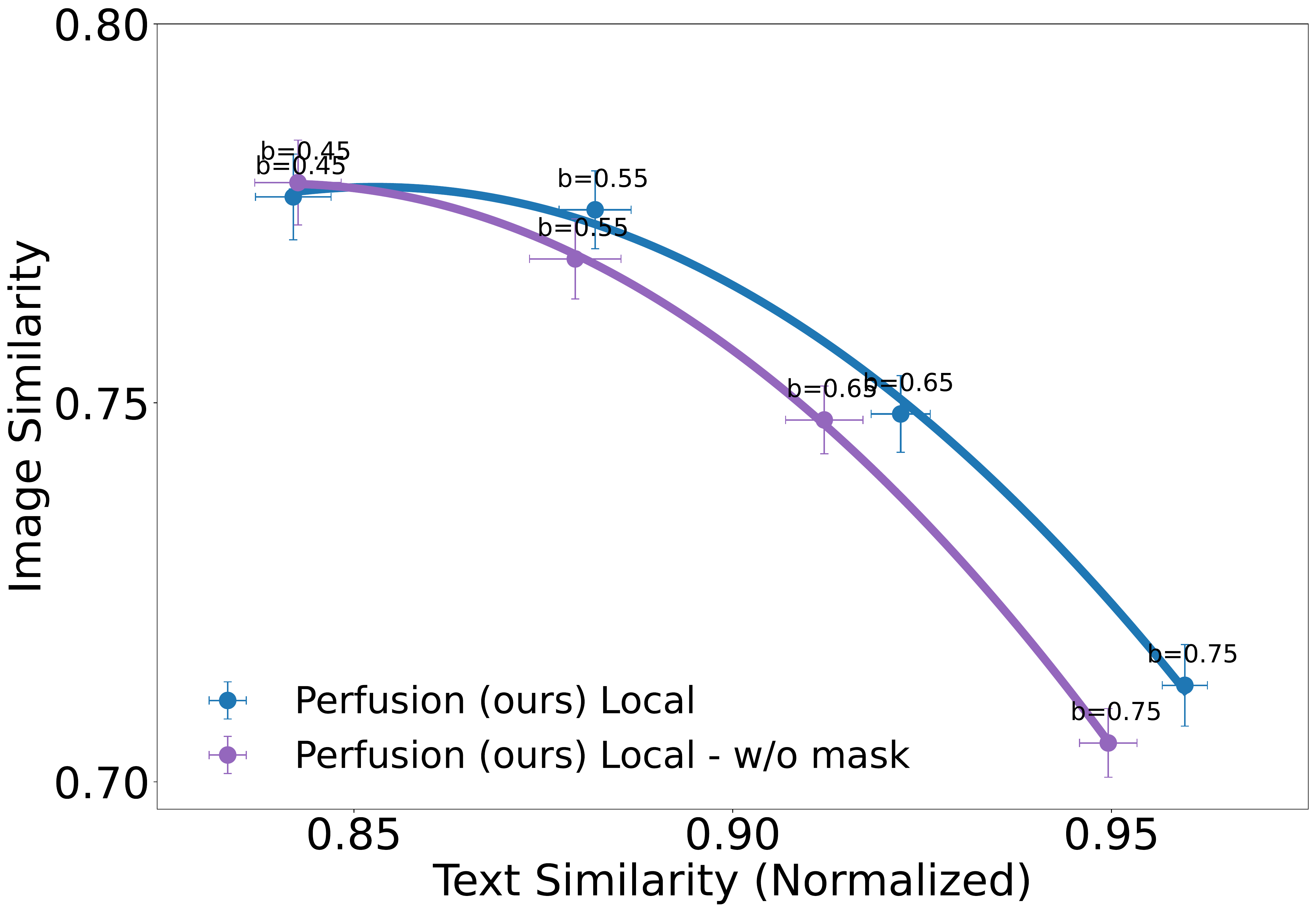} %
    \put (90,60) {(c)}
    \end{overpic}
\begin{overpic}[width=0.49\columnwidth, trim={0.cm 0cm .32cm .3cm},clip]{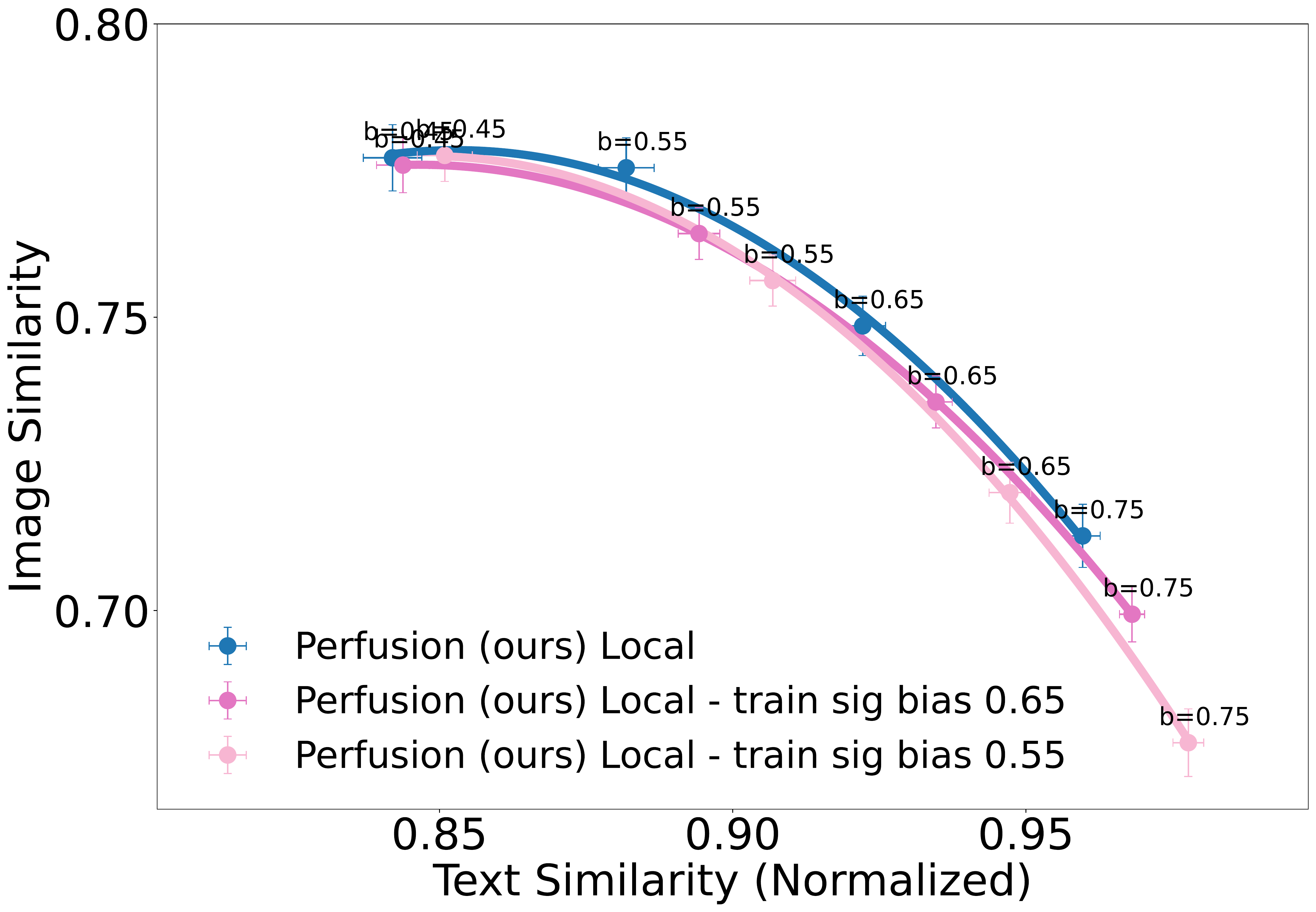} %
    \put (90,60) {(d)}
    \end{overpic}

\begin{overpic}[width=0.49\columnwidth, trim={0.cm 0cm .32cm .3cm},clip]{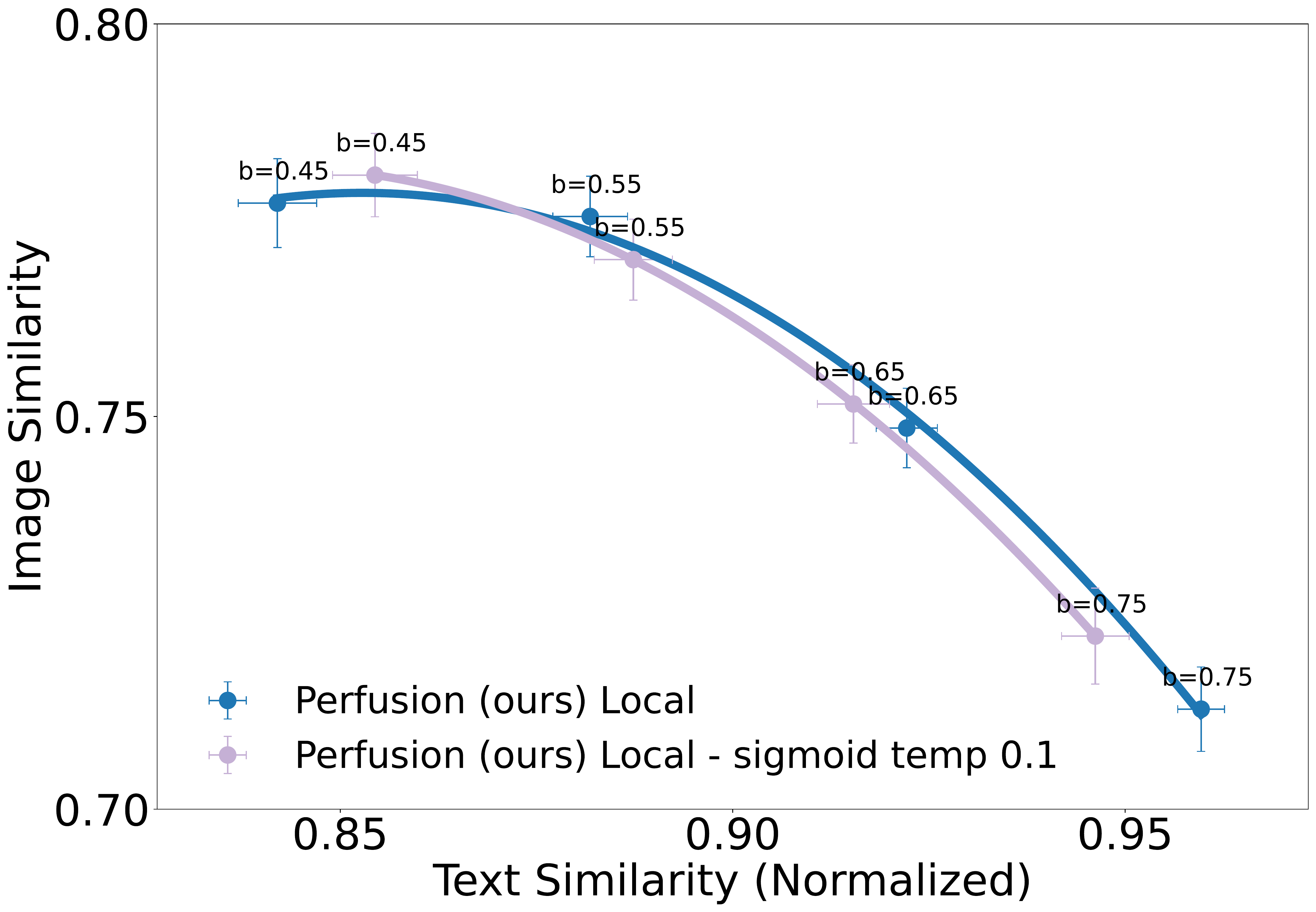} %
    \put (90,60) {(e)}
    \end{overpic}
\hspace{0.49\columnwidth}
\caption{ \textbf{Ablation study:} We show the Pareto front for various ablation conditions. \textbf{(a) 1-shot training:} our method trained with a single image per concept. \textbf{(b) Trained K:} We train the key pathway as well, instead of locking it. \textbf{(c) Usage of Zero-shot Mask During Training:} We compare \ourmethod with and without the zero-shot mask. \textbf{(d) Sigmoid Bias During Training:} We compare training with different values of sigmoid bias (our default value is 0.75). \textbf{(e) Sigmoid Temperature During Inference:} We compare inference with sigmoid temperature of 0.1 to the default temperature of 0.15.}
\label{fig_ablation_all}
\vspace{-5pt}
\end{figure}

\subsection{User Study}
We further evaluate the models through two user studies conducted with Amazon Mechanical Turk. In the first study, raters were given two images of a concept and a prompt. They were asked to rank images generated by the three methods (\ourmethod, CD, DB), based on how well they portrayed the concept according to the prompt. We used $11$ concepts, $24$ prompts per concept, with $8$ responses per prompt. \ourmethod was selected first with an average rank of
$2.18 \pm 0.02$ (SEM), CD was 2nd ($2.06 \pm 0.02$) and DB was last ($1.75 \pm 0.02$), demonstrating a preference for our approach.
For the second study, we investigated whether \ourmethod harms the generative prior. To do so, we compared \ourmethod to ``vanilla'' stable-diffusion (SD). Raters were shown a prompt and two generated images, one by \ourmethod and another by SD. They were asked to rate the images according to their realism, using a score between $1$ to $3$ (best). We gathered the same number of concepts, prompts and responses as the first study. \ourmethod had an average score of $1.885 \pm 0.017$ (SEM), while SD had a score of $1.894 \pm 0.017$. The results are statistically indistinguishable, demonstrating that \ourmethod can preserve the generative prior.
See Appendix \ref{suppl_user_study} for additional details on these experiments.
\color{black}

\subsection{Ablation study}
\label{sec_supp_ablation}

In \figref{fig_ablation_all}  we study in greater depth the properties of \ourmethod by an ablation study. We show the trade-off between visual and textual similarity for the following conditions.

\begin{enumerate}[label=(\alph*)]
    \item \textbf{1-shot:} We compare between training our method with all the training examples (average of $6.5$ image for each concept), to training with just a single example for each concept. We observe that training with just a single example introduces slight overfitting.%
    \item \textbf{Key-locking:} We compare between our method with key-locking to our method with trained key projection layers. It is evident that key-locking shifts the Pareto curve to the right - meaning less overfit. This result confirms our hypothesis that locking the key projection layers leads to better textual-alignment and enables complex deformations of the learned concept.
    \item \textbf{Zero-Shot Masking Loss:} We compare the effects of training with and without a zero-shot mask. We notice that using zero-shot mask tends to improve the textual similarity, which mean it helps reduce the overfitting.
    \item \textbf{Sigmoid Train Bias}: We compare between different values of the Sigmoid biases used during training time. We notice that using a higher bias results in better Pareto front. 
    \item \textbf{Sigmoid Inference Temperature} We compare between different values of the Sigmoid temperature used during inference time. We notice that using inference-time Sigmoid temperature that is higher than the train-time Sigmoid temperature results a better pareto front.  Generally temperature of 0.15 tends to work better.

\end{enumerate}

\begin{figure*}[t]
\vspace{-0pt}
\includegraphics[width=\textwidth, trim={0.cm 20.5cm 0.cm 0.cm},clip]{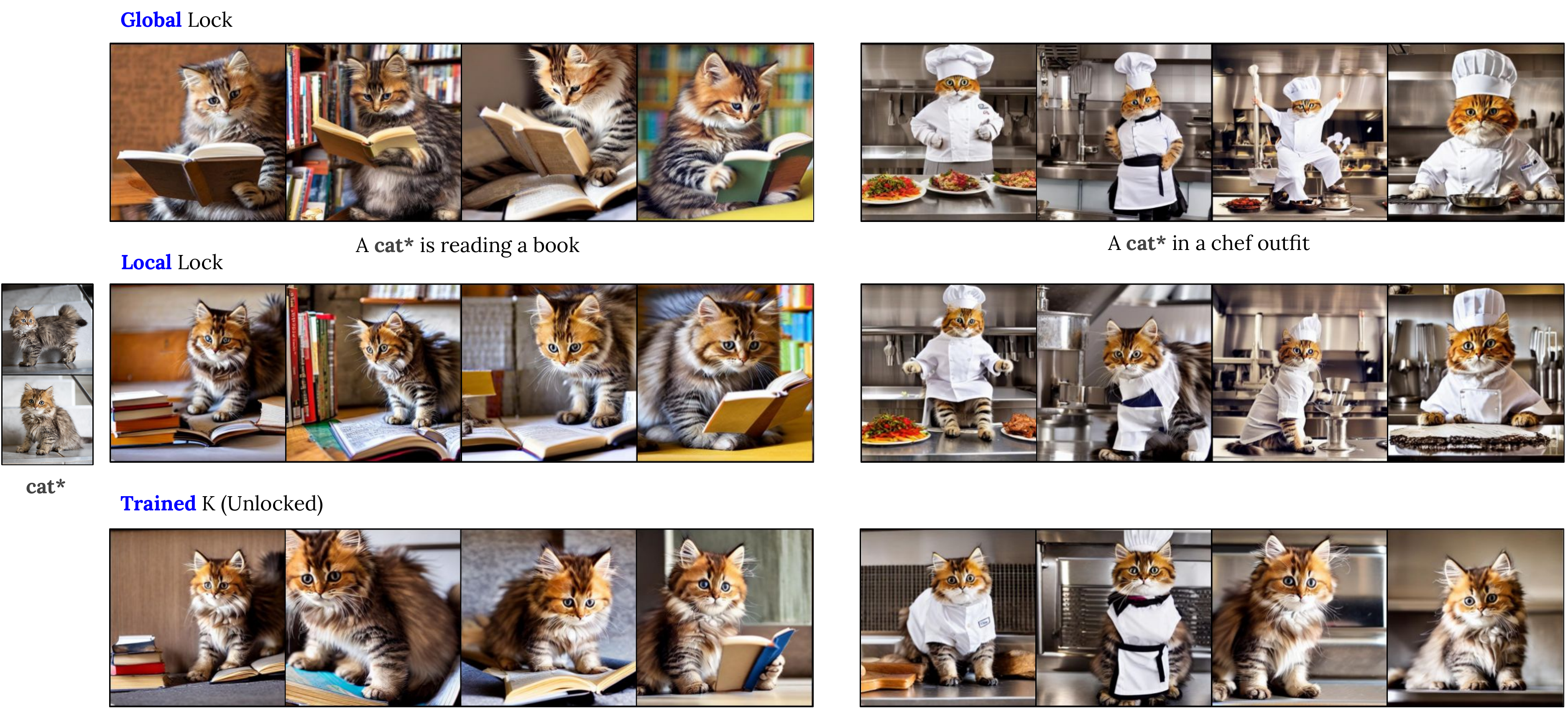} %
    \caption{ \textbf{Comparing lock types:} \textit{Global key-locking} allows for more visual variability and can accurately portray the nuances of an object or activity, like when depicting the cat \textit{in a human-like posture} reading a book or wearing a chef outfit. \textit{Local key-locking} also has successes, but they are not as effective as global key-locking. Finally, \textit{Trained-K} has better compatibility with the training images, but it sacrifices its alignment with the text.
}
\label{fig_lock_types}
\vspace{-5pt}
\end{figure*}

\begin{figure*}[t]
\vspace{-0pt}
\includegraphics[width=\textwidth, trim={0.cm 25cm 0.7cm 0.3cm},clip]{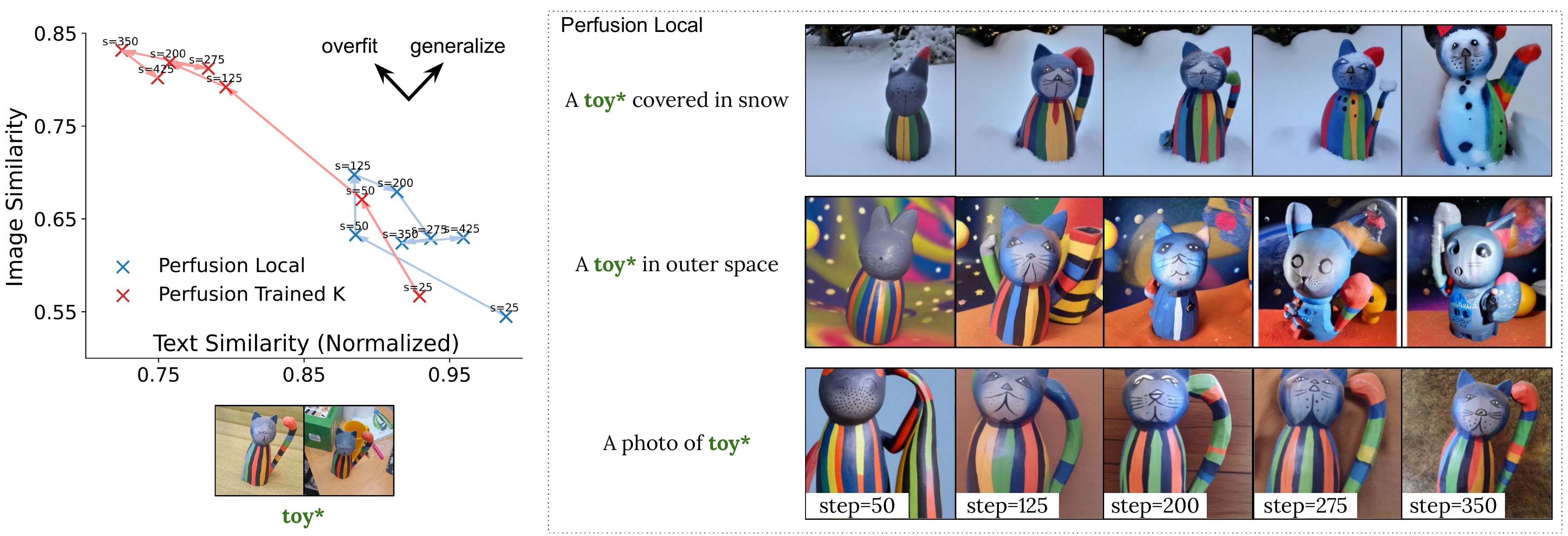} %
    \caption{ \textbf{The impact of Key-Locking on training dynamics:}
Comparing \textit{Trained K} (without lock) and \textit{Local Key-Locked}  on the plane of fidelity vs. textual alignment, where `s'' denotes the training step. \textit{Trained K} overfits the training images, but negatively affects textual alignment. Conversely, \textit{Local Key-Locked} training exhibits ``over-generalization'', leading to improved textual alignment, but reduced visual fidelity.
As training progresses, qualitative examples reveal that the model becomes increasingly toy-like. This suggest that the V features learn latent properties of the supercategory to improve the alignment between the Q and K features.
Finally, the cat-toy in this example demonstrates a failure case as the V features pick up the toy supercategory too early in the training.
}
\label{fig_lock_curves}
\vspace{-5pt}
\end{figure*}

\section{Qualitative visual comparisons}
Next, we provide qualitative comparisons that reveals the strengths of our approach, \edit{along with examples of our main failure mode.}

\myparagraph{Single Concept Text-guided Synthesis:} Figures \ref{fig1}, \ref{fig_single} and \ref{fig_single_extra_page} show our ability to compose novel scenes when using \ourmethod and compare them with strong baselines. For each concept, we show exemplars from our training set, along with generated images and their conditioning texts. Our approach allows making deformations to the concept appearance without losing its identity. At the same time, it correctly encapsulates the semantic qualities of both the concept and the prompt.
For example, notice how we can fully customize the teddy* concept in \figref{fig1} and the cat* in \figref{fig_single} with different garments, without compromising their identities, and at the same time allow them to interact with the scene. We can also change the material of the teapot* to pure gold or transparent glass, while retaining its distinctiveness. In Fig. \ref{fig_single} \ourmethod is the only one that can shatter the pot* concept. Notice how it can change the dog* posture, making it appear as though it is engaged in reading a book, with its eyes focused on the text and its paws  grasping the book. \edit{In Fig. \ref{fig_single_with_TI} we provide additional results including Textual Inversion.}

\myparagraph{Multi Concept Text-guided Synthesis:} Figures \ref{fig1} and \ref{fig_multi} show our ability to compose novel scenes with multiple concepts, when using \ourmethod and compare them with CD's ``optimization'' approach to combine individual concepts. We use their provided images when comparing with prompts from their paper, otherwise with use ``Better-HP". Notice how \ourmethod can compose the teddy* and teapot* in different scenes, or how it allows the teapot* to hold the teddy* while sailing. When comparing with CD, we observe similar or better results. For example, observe how the water-color painting better preserve the chair identity, or how the teddy* can wear the sunglasses* successfully.

\myparagraph{Balancing visual-fidelity and textual-alignment:} \figref{fig_bias} provides qualitative examples for balancing the visual-fidelity versus the textual-alignment, by adjusting the sigmoid bias threshold. Higher bias values reduce the impact of the concept, while lower values give it more prominence in the generated image. This is because the concept energy is spread across multiple encodings in the text encoder, not just the one corresponding to the concept word. Lowering the bias increases its influence on all relevant encodings.

Next, we demonstrate several aspects of the key-locking mechanism. We start by comparing global key-locking local key-locking and no key-locking. Then, we show what happens when we lock concepts to different super-categories. Finally, we study the training dynamics of local key-locking and show an ``over-generalization'' phenomena that makes it over-align with the supercategory.

\subsection{Key Design Decisions}
Next, we demonstrate and discuss key design decisions. We start by comparing global key-locking local key-locking and no key-locking. Then, we show what happens when using vanilla ROME, when there a train-inference mismatch, and the weight update is performed only after the optimization step.

\myparagraph{Comparing lock types:}
Fig. \ref{fig_lock_types} compares \textit{global key-locking}, \textit{local key-locking} and \textit{trained-K} (no key-locking, K pathway is trained like the V pathway). We find that global locking allows to generate rich scenes, portray better the nuances of the object attributes or activities, and in general allow more visual variability of the concept, compared to local key locking and trained-K. For example notice the cat depicted in human-like postures while reading a book, or when wearing a chef outfit. Local locking also has some successes but they are weaker. Finally, trained-K is more aligned with the postures and appearance of the training images, while sacrificing alignment with the text.

\noindent\myparagraph{Train-inference mismatch, when using vanilla ROME:}
Fig. \ref{fig_missmatch} illustrates the mismatch between generated images during training and inference when editing with vanilla ROME. This results in corrupted images during inference.

\begin{figure}[h]
\vspace{-0pt}
\includegraphics[width=\columnwidth, trim={1.cm 10.cm 13.5cm 1.cm},clip]{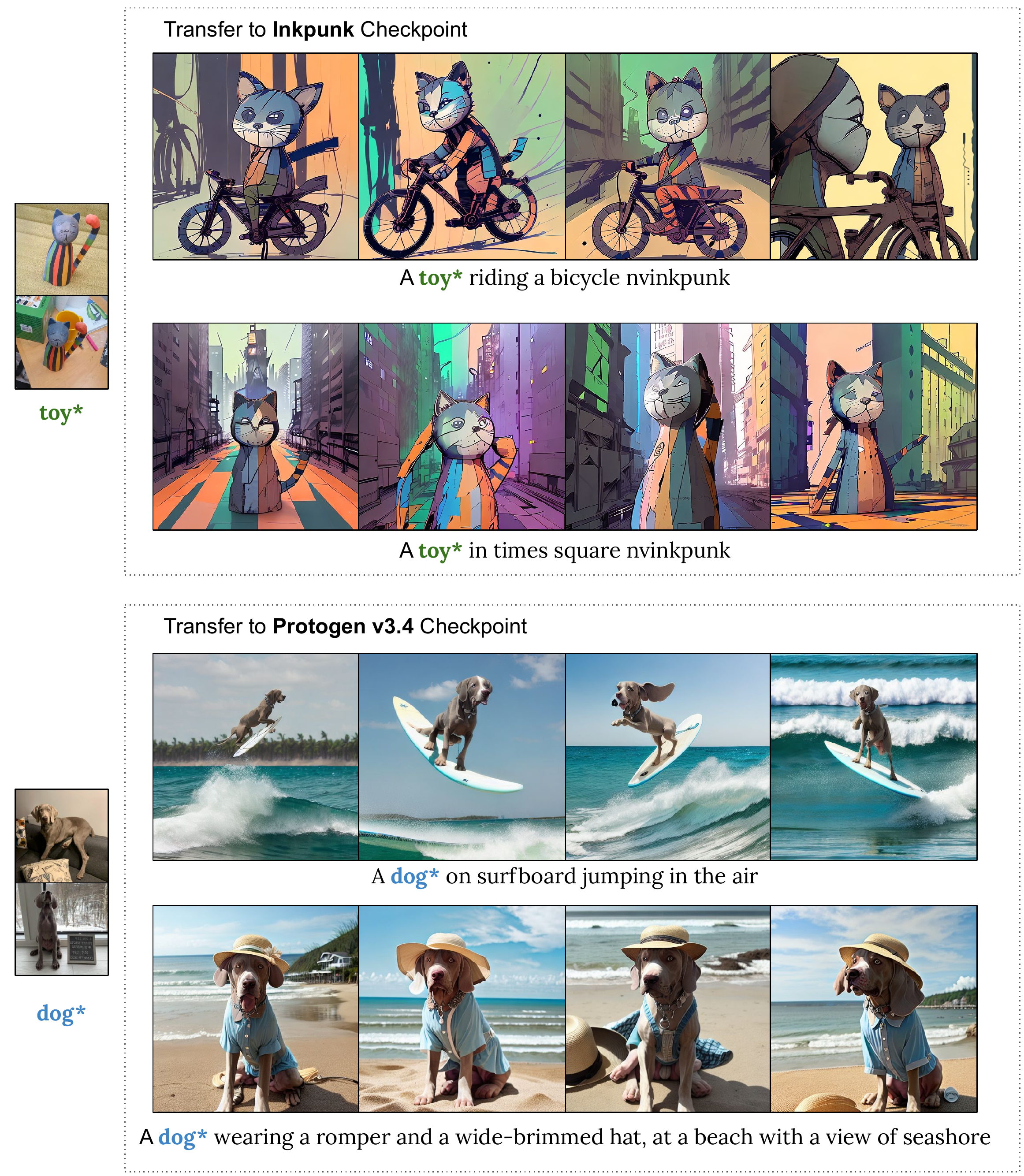} %
    \caption{ \textbf{Zero-shot transfer to fine-tuned models:} A Perfusion concept trained using a vanilla diffusion-model can generalize to fine-tuned variants.
}
\label{fig_zs_ckpt}
\vspace{-5pt}
\end{figure}

\begin{figure}[h]
\vspace{-0pt}
\includegraphics[width=\columnwidth, trim={3.4cm 19.5cm 7.8cm 4.6cm},clip]{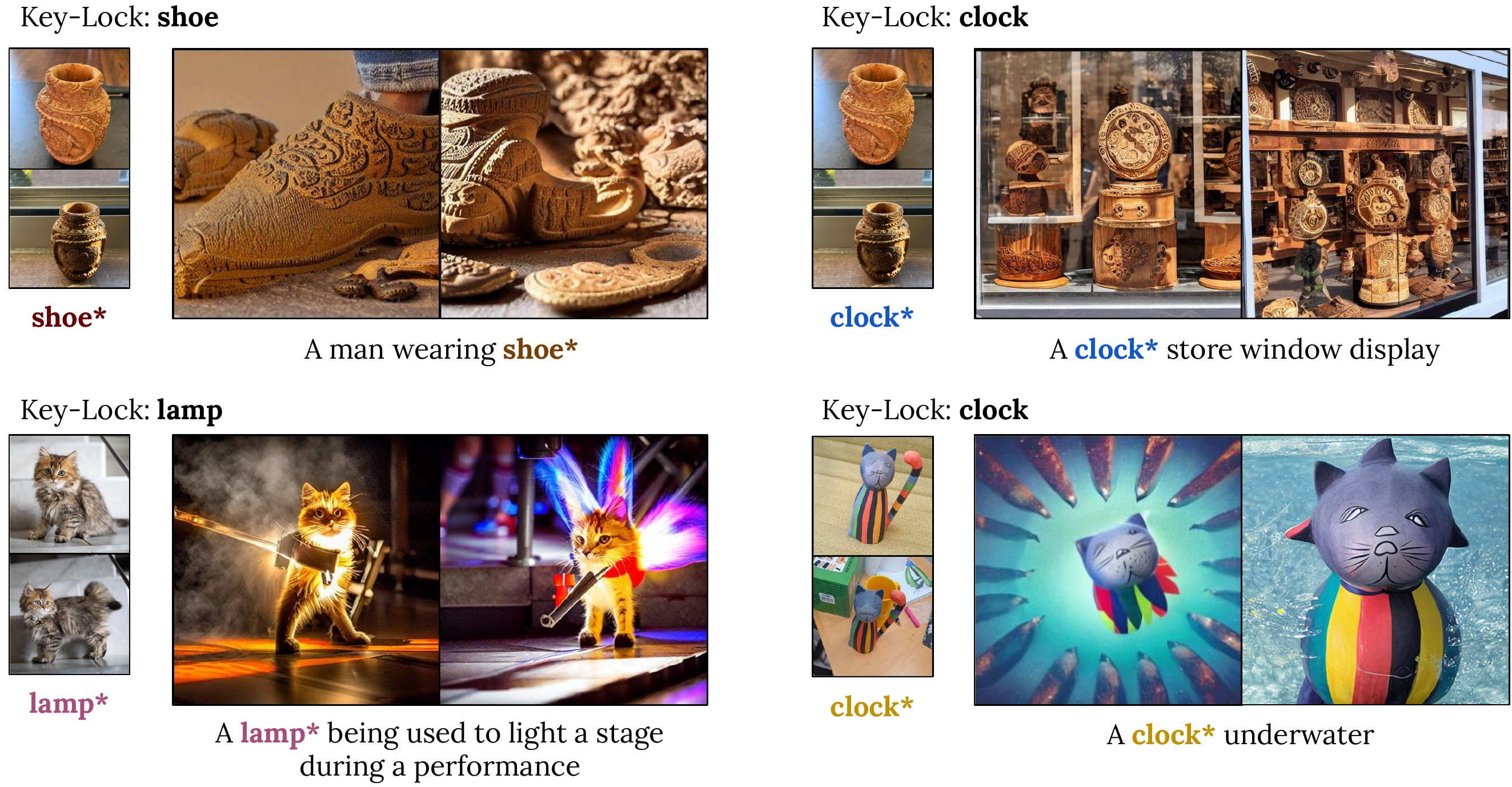} %
    \caption{ \textbf{Locking to unusual super-categories:} A concept ``inherits'' the qualitative outline of the supercategory it is locked onto. For example, the pot as a shoe or a clock, and cat becomes illuminated when locked to lamp.}
\label{fig_diverse_locks}
\vspace{-5pt}
\end{figure}

\subsection{Robustness Analysis}
Next, we will show how our approach performs in different situations, highlighting both its strengths and weaknesses.

\myparagraph{One-shot learning:}
We compare training our method with all examples (average of 6.5 images/concept) to using only one example per concept. In Figures \ref{fig_oneshot_qualitative} and \ref{fig_ablation_all}(a), we note a slight overfit when training with only one example.

\myparagraph{Zero-shot transfer to fine-tuned checkpoints:}
A Perfusion concept trained using a vanilla diffusion-model could generalize to fine-tuned variants of the model. Fig. \ref{fig_zs_ckpt} shows transfer abilities to two popular variants of Stable-Diffusion: InkPunk-v2 \cite{inkpunkv2} and Protogen-v3.4 \cite{protogen34}.
\color{black}

\begin{figure*}[t]
    \centering
    \includegraphics[width=\textwidth, trim={0.cm 0.cm 0cm 0cm},clip]{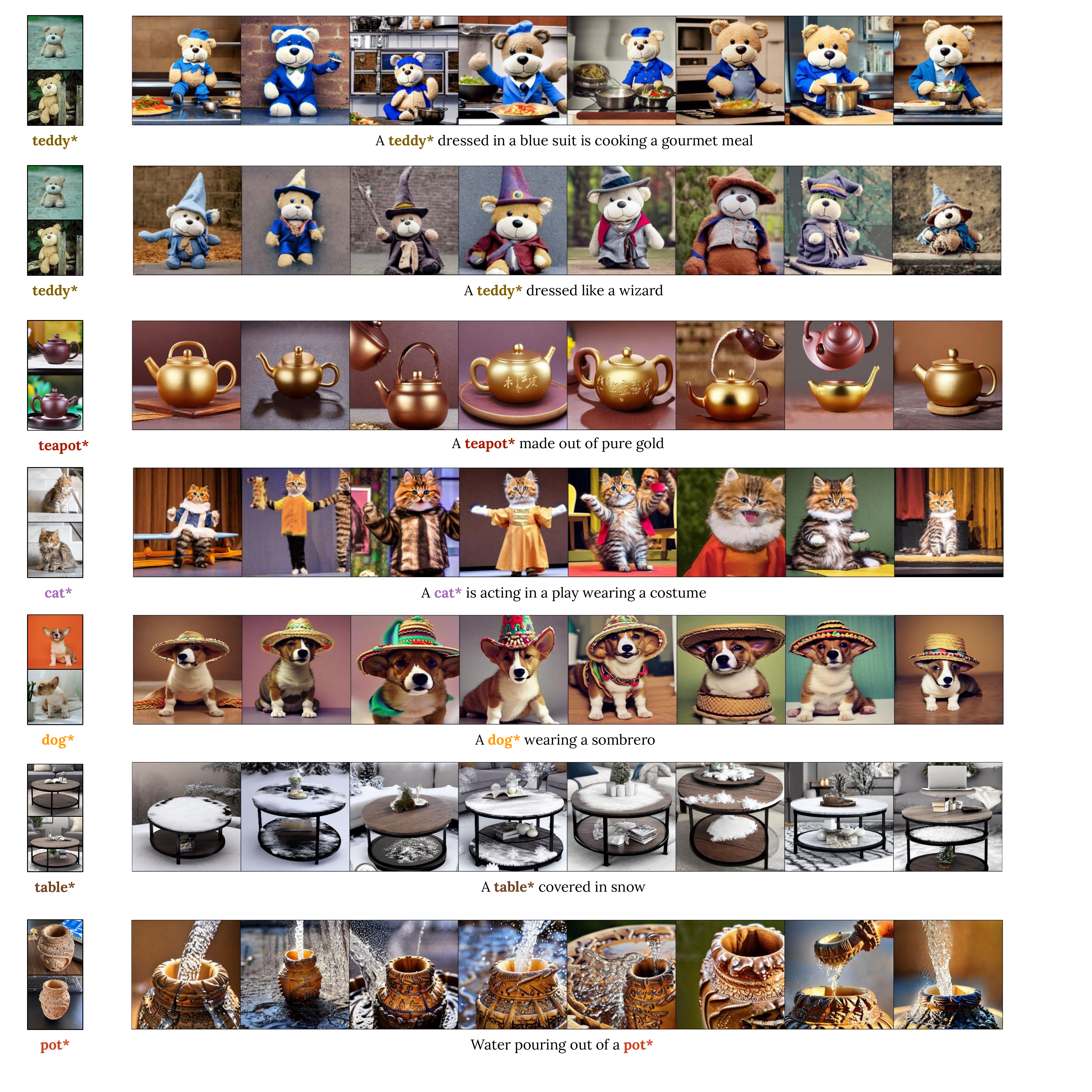} %
    \caption{\textbf{Uncurated samples of image variations with text guided prompts.} We observe that a batch size of 8 is typically sufficient to ensure several good samples.}
    \label{fig_uncurated_extra_page}
\end{figure*}

\begin{figure*}[h]
\centering
    \includegraphics[width=\textwidth, trim={0.cm 0.4cm 3cm .5cm},clip]{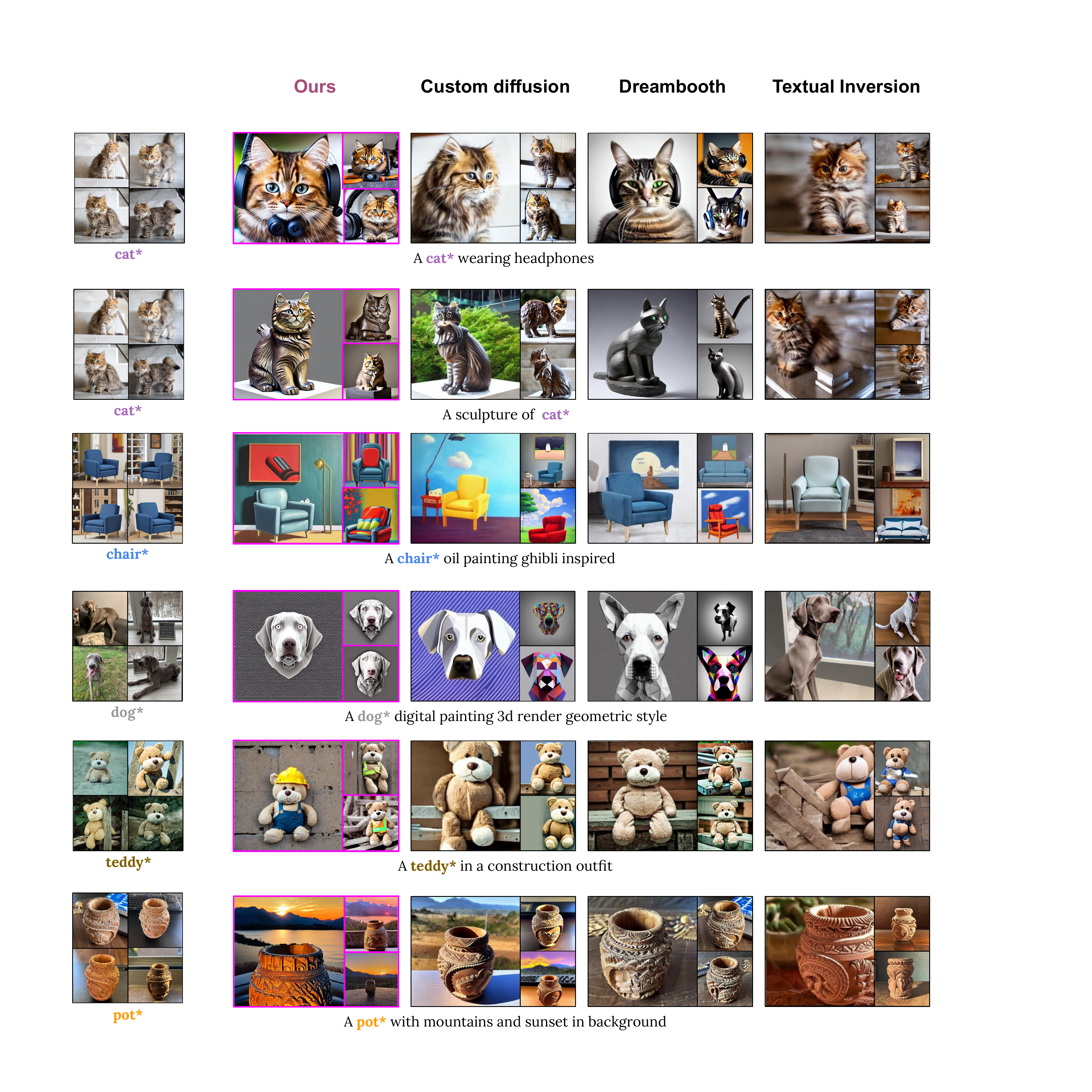}  %
    \caption{\textbf{Supplement generation results with single concept examples.} For each concept, we show exemplars from our training set, along with generated images,  their conditioning texts and comparisons to Custom-Diffusion (CD),  Dreambooth (DB), \edit{and Textual-Inversion (TI)} baselines. All prompts are taken from the CD paper, as well as the baseline generations.}
    \label{fig_single_extra_page}
\end{figure*}

\myparagraph{Uncurated samples:} Fig. \ref{fig_uncurated_extra_page} shows that a batch size of 8 is typically sufficient to ensure several good samples.

\myparagraph{Locking to unusual super-categories} \figref{fig_diverse_locks} shows how the concepts are portrayed when locked to unusual super-categories. We observe that the concepts ``inherit'' the qualitative outline of the unusual super-category. When the pot is locked to a shoe or a clock, it is portrayed in the outline of a shoe or clock, but in the style of the pot. When the cat is locked to a lamp, it becomes illuminated.

\myparagraph{The impact of Key-Locking on training dynamics:}
The left panel of \figref{fig_lock_curves} shows the comparison between \textit{Trained K} and \textit{Local} Key-Locked training for various training steps (``s'').
As training progresses, \textit{Trained K} (no-lock) overfits the training images, while hurting the textual alignment. Interestingly, Local Key-Locked training reveals an ``over-generalization'' phenomenon. As training progresses the concept learns supercategory (latent) features, improving textual alignment but sacrificing visual fidelity.
The right panel of \figref{fig_lock_curves} displays qualitative examples, where longer training makes the concept ``in outer space'' increasingly toy-like. It is worth noting that reconstruction prompts have better visual fidelity as the concept is trained using such prompts. This suggests that the learned supercategory features are latent.
These findings were expected, as the learned V features propagate to the Q features in the next denoising step and the Q features should align with the locked supercategory K features due to their inner product when calculating the attention map. Hence, we expected that the V features should learn to encode latent properties of the supercategory in order to improve Q-K alignment.
Finally, the example of the cat-toy demonstrates a failure case where the V features pick up the supercategory features too early during training, before the concept has completed learning its fine-grained characteristics.

\section{Conclusions and Limitations}
We have presented \textit{\ourmethod}, a novel \TTI personalization method that combines high visual fidelity with improved textual alignment. Our approach, which utilizes a gated rank-1 method, provides control over the influence of learned concepts during inference time, enables combination of multiple concepts, and results in a small model size. Importantly, the key-locking technique leads to novel qualitative results compared to traditional approaches.

 \myparagraph{Limitations and future work.}
 We find that the choice of supercategory word to lock onto may sometime produce ``over-generalization'' effects when using the concept in a new prompt. \edit{For instance as shown in \figref{fig_lock_curves},} setting the concept of a toy-cat as a ``toy'' may encourage learning to generate it in a childish style in new prompts, while scarifying its visual fidelity. \edit{Additionally, \figref{fig_diverse_locks} demonstrates that locking concepts to atypical super-categories results in the concepts adopting some characteristics of that atypical super-category.}
 A second limitation is that combining concepts requires a great deal of prompt engineering. Interestingly, we found it was easier to succeed on prompts that were suggested by CD, than with prompts that we devised. We believe that there is much of a headroom for improvement in this task.

\begin{acks}
We are grateful to the anonymous reviewers of SIGGRAPH 2023 for their valuable feedback, which greatly improved the final version. Additionally, we would like to thank Assaf Hallak, Xun Huang, Jim Fan, and Eli Meirom for providing insightful feedback on an earlier draft of this manuscript. This work was carried out in partial fulfillment of the requirements for the Ph.D. degree of the first author.
\end{acks}

\clearpage

\clearpage

\clearpage

\bibliographystyle{ACM-Reference-Format}
\bibliography{main}

\clearpage

\appendix

\renewcommand{\thefigure}{A.\arabic{figure}}
\setcounter{figure}{0}

\textbf{\huge Appendix: Key-Locked Rank One Editing for Text-to-Image Personalization}

\section{Lemma: Derivation of the Expression for the Weight Update of ROME}
\label{sec_supp_lemma}

In this section we provide a Lemma that reorganizes and calculates the output of a layer affected by equation \eqref{eq_ROME} repeated below for convenience.

\begin{align*}
    \hat{W} = W + \Lambda (C^{-1}\istar)^T,
\end{align*}
where $\Lambda = (\ostar - W \istar)/[(\istar^T (C^{-1})^T \istar)]$, $C$ is a constant positive definite matrix, that is pre-cached.

\textbf{Lemma:}
Given an input \Em, the output $h$ of an edited layer is:
\begin{align}
        h = W \Em^\perp + {\ostar sim(\istar, \Em)}/{||\istar||^2_{C^{-1}}},
    \label{eq_ROME_sim_appendix}
\end{align}
where (1) $sim(\istar, \Em) := \istar^T (C^{-1})^T \Em$ measures the similarity of \Em with $\istar$ in a metric space defined by $C^{-1}$, (2) $||\istar||^2_{C^{-1}} := sim(\istar, \istar)$ measures the energy of $\istar$ in the same metric space. (3) $\Em^\perp := \left(\Em - {\istar \textit{sim}(\istar, \Em) }/{||\istar||^2_{C^{-1}}}\right)$ is the component of $\Em$ that is orthogonal to $\istar$ in the metric space.

\myparagraph{Proof:}

 We now proceed to prove the Lemma.

Let $h$ be the output of the edited layer given an input $\Em$. From equation \eqref{eq_ROME}, we have
\begin{align*}
h = \hat{W}\Em = W\Em + \Lambda (C^{-1}\istar)^T\Em.
\end{align*}

Notice that the term $(C^{-1}\istar)^T\Em$ is the similarity of $\Em$ with $\istar$ in the metric space defined by $C^{-1}$, denoted by $sim(\istar, \Em) =: \istar^T (C^{-1})^T \Em$. Thus, we can rewrite the above equation as
\begin{align*}
h = W\Em + {\Lambda \textit{sim}(\istar, \Em)}.
\end{align*}

To simplify the above expression, we examine the term $\Lambda$. Recall that $\Lambda = (\ostar - W \istar)/[(\istar^T (C^{-1})^T \istar)]$. Thus, we can substitute and rearrange the above equation to obtain
\begin{align*}
h &= W\Em + {\frac{\ostar - W \istar}{\istar^T (C^{-1})^T \istar}\textit{sim}(\istar, \Em)} \\
&= W\Em + {\frac{\ostar}{\istar^T (C^{-1})^T \istar}\textit{sim}(\istar, \Em)} - {\frac{W \istar}{\istar^T (C^{-1})^T \istar}\textit{sim}(\istar, \Em)}.
\end{align*}

We now define $||\istar||^2_{C^{-1}} = \istar^T (C^{-1})^T \istar$ for brevity. With this, we can rewrite the above equation as
\begin{align*}
h = W\Em + {\frac{\ostar}{||\istar||^2_{C^{-1}}}\textit{sim}(\istar, \Em)} - {\frac{W}{||\istar||^2_{C^{-1}}} \istar\textit{sim}(\istar, \Em)}.
\end{align*}

To further simplify the above expression, we define $\Em^\perp = \\ \left(\Em -  {\istar \textit{sim}(\istar, \Em) }/{||\istar||^2_{C^{-1}}}\right)$. This is the component of $\Em$ that is orthogonal to $\istar$ in the metric space defined by $C^{-1}$. With this, we can rewrite the above equation as
\begin{align*}
h &= W\Em^\perp + {\frac{\ostar}{||\istar||^2_{C^{-1}}}\textit{sim}(\istar, \Em)},
\end{align*}
as stated in the Lemma.

Namely, the left term nulls the ``energy'' of $\istar$ from \Em and passes the rest ($\Em^\perp$) through the original matrix $W$. Then, the right term grossly assigns that energy in the direction of the $\ostar$ component.

\section{Derivation of $\mathbf{e}_m^{\perp J}$, the orthogonal component for multiple concepts}
\label{supp_span_orthogonal}

To calculate the orthogonal component of \Em for multiple concepts, we start by projecting \Em and the set of target-inputs $\{\istar^j\}_{j=1..J}$ to the metric space of $C^{-1}$, using a Cholesky decomposition $C^{-1} = L L^T$
\begin{align}
\widetilde{\Em} & = L^T \Em \nonumber \\
\widetilde{\istar^j} & = L^T \istar^j,
\end{align}
where we denote by ``$\sim$'' every vector in the metric space.

Next, we use a QR decomposition to find an orthonormal spanning basis for
$\{\widetilde{\istar^j}\}_{j=1..J}$. We denote this basis by $\{\widetilde{u^j}\}_{j=1..J}$.

Then, the orthogonal component of $\widetilde{\Em}$ in this metric space is

\begin{align}
    \widetilde{\Em^{\perp J}} = \widetilde{\Em} - \sum_{j \in 1..J}{\widetilde{u_j} \braket{\widetilde{u_j},\widetilde{\Em}} },
\end{align}
where $\braket{\widetilde{u_j},\widetilde{\Em}}$ is simply their dot product $\widetilde{u_j}^T \cdot \widetilde{\Em}$.

Projecting the last expression back to text encoder space using the inverse of the Cholesky root ${L^T}^{-1}$, mutiplied from the left yields

\begin{align}
    {\Em^{\perp J}} = {\Em} - \sum_{j \in 1..J}{u_j \braket{\widetilde{u_j},\widetilde{\Em}} },
\end{align}
where $u_j = {L^T}^{-1}\widetilde{u_j}$.

Finally, note that $\braket{\widetilde{u_j},\widetilde{\Em}} = sim(u_j, \Em)$, because

\begin{align}
    \braket{\widetilde{u_j},\widetilde{\Em}} = \widetilde{u_j}^T \cdot \widetilde{\Em} = u_j L L^T \Em = sim(u_j, \Em)
\end{align}

 Which brings us to our final expression for $\Em^{\perp J}$

 \begin{align}
  \Em^{\perp J} = \Em - \sum_{j \in 1..J}{u_j\textit{sim}(u_j, \Em) }
  \label{eq_Em_perp_J_supp}
 \end{align}

\subsection{The special case of a single concept}
\label{supp_single_orthogonal}

As a final step, we prove that when there is only a single concept, denoted by $J=1$, then $\Em^{\perp J} = \left(\Em - {\istar \textit{sim}(\istar, \Em) }/{||\istar||^2_{C^{-1}}}\right)$.

By definition, $\widetilde{u_1} = L^T \istar / ||L^T \istar||$, and thus $u_1 = \istar / ||L^T \istar||$. Substituting this into Equation \eqref{eq_Em_perp_J_supp}, we obtain:

\begin{align}
\Em^{\perp J} = \Em - \frac{\istar}{||L^T \istar||} \textit{sim}(\frac{\istar}{||L^T \istar||}, \Em) = \Em - \frac{\istar \textit{sim}(\istar, \Em)}{||L^T \istar||^2}
\end{align}

To complete the proof, we note that $||L^T \istar||^2 = \istar^T C^{-1} \istar = ||\istar||^2_{C^{-1}}$, by definition.

This concludes the proof that $\Em^{\perp J} = \left(\Em - {\istar \textit{sim}(\istar, \Em) }/{||\istar||^2_{C^{-1}}}\right)$ when there is only a single concept, $J=1$.

\begin{figure*}[h]
    \includegraphics[width=0.95\textwidth, trim={0cm 23.5cm 0cm 0.3cm},clip]{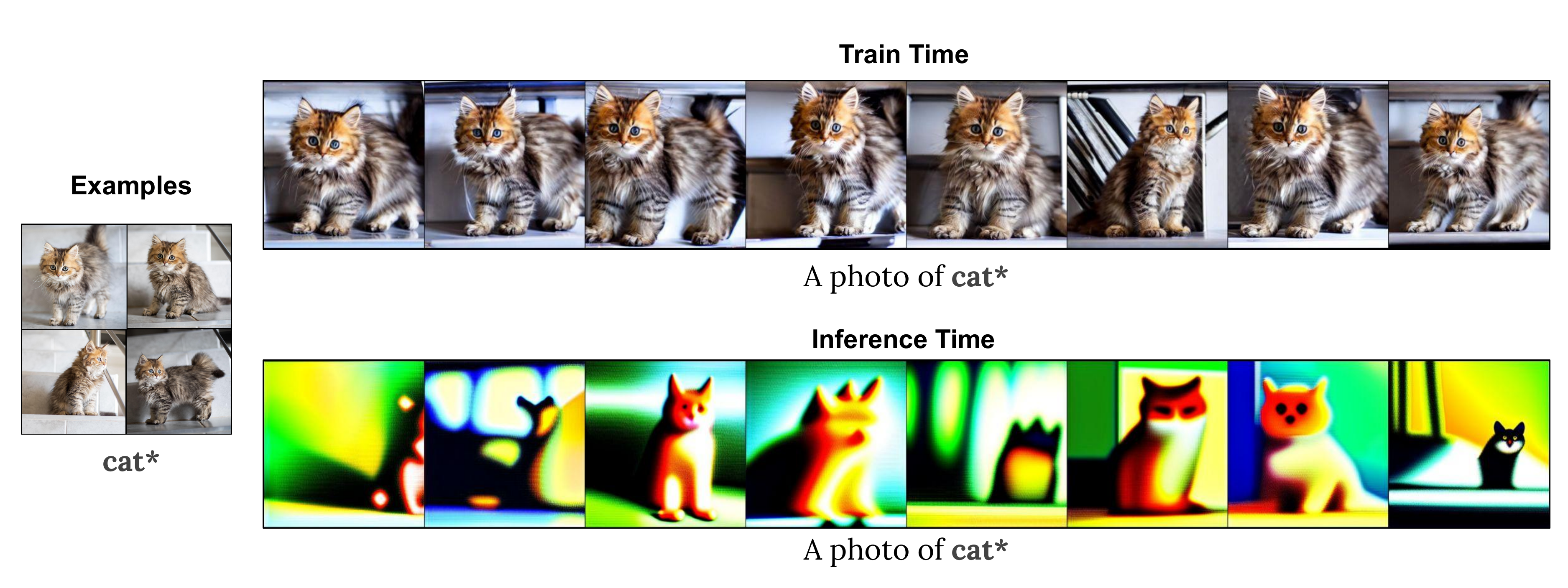} %
    \vspace{-10pt}
    \caption{
    \edit{\textbf{Train-Inference Mismatch:} As presented in Challenge 1, when training the model with ROME's naive solution, we observe corrupted generations during inference time. We show that the low quality of generated images is the result of the mismatch between the objective during inference and training time.}
    }
    \label{fig_missmatch}
\end{figure*}

\begin{figure*}[h]
\centering
    \includegraphics[width=\textwidth, trim={0.cm 8.cm 3cm .5cm},clip]{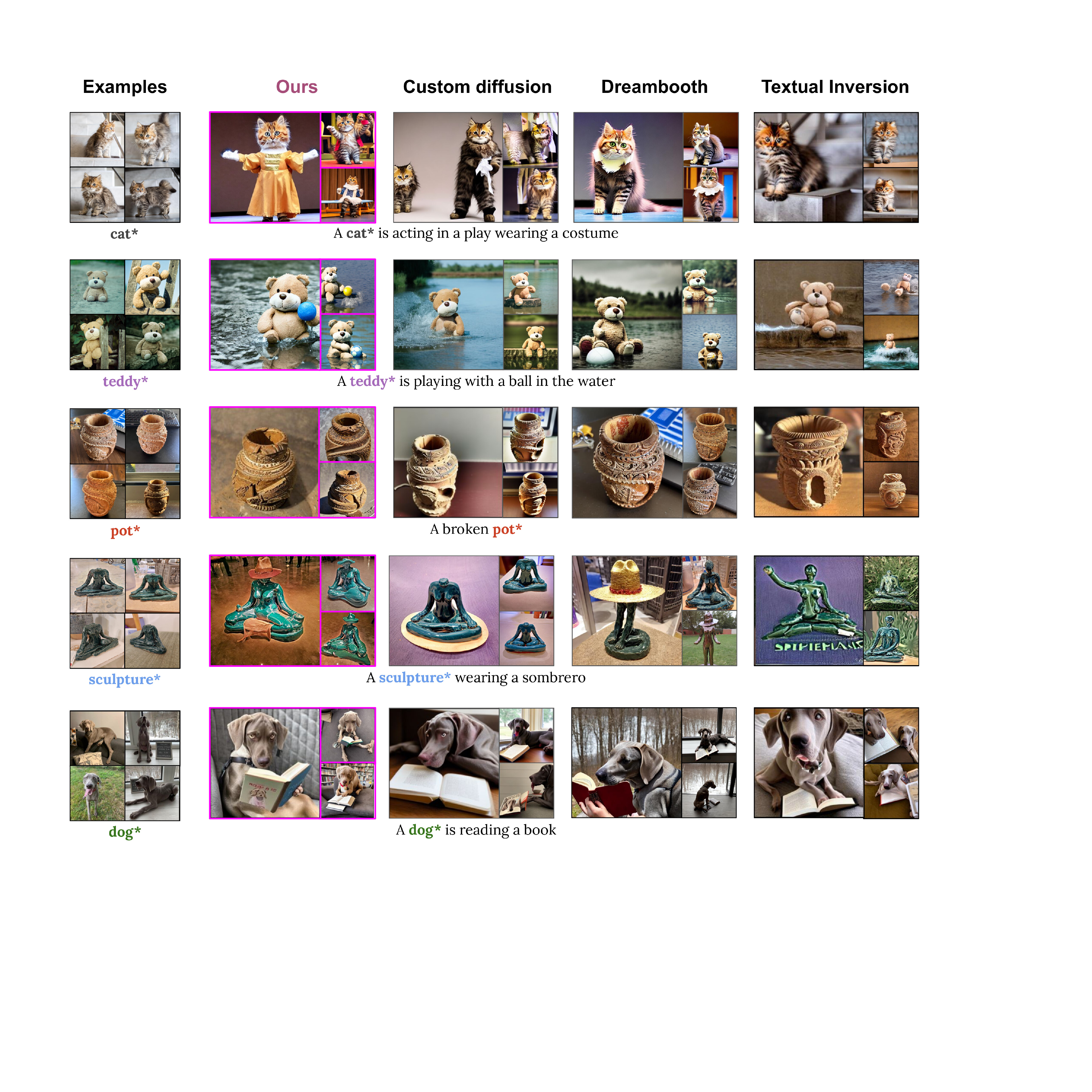}  %
    \caption{\edit{\textbf{Qualitative comparison including Textual Inversion.}}}
    \label{fig_single_with_TI}
\end{figure*}

\begin{figure*}[h]
\centering
    \includegraphics[width=\textwidth, trim={4.cm 22cm 11cm .5cm},clip]{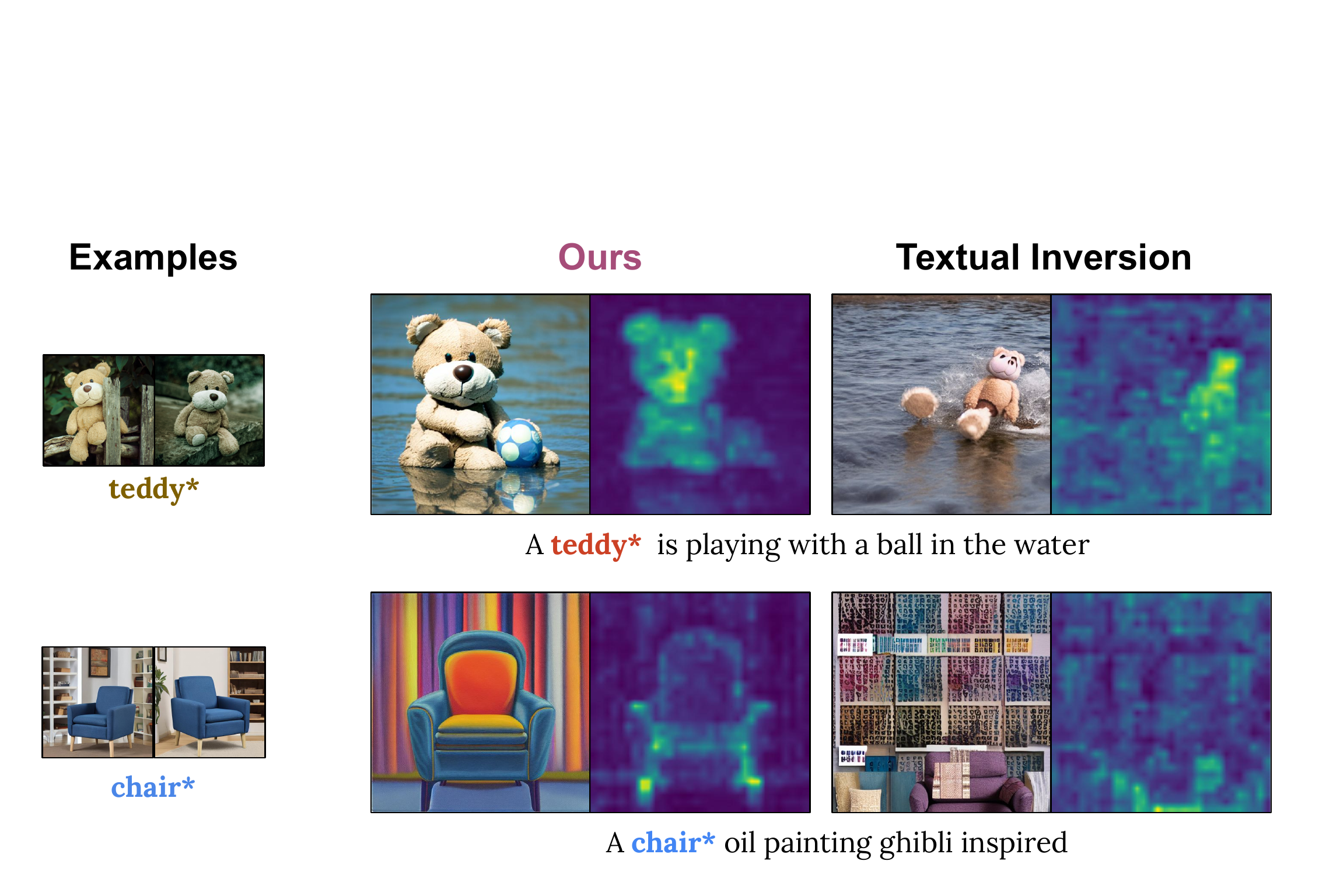}  %
    \caption{\edit{\textbf{Additional results of attention overfit in Textual inversion}}}
    \label{fig_attention_appendix}
\end{figure*}

\section{Additional technical details}
\label{sec_supp_additional_details}
\myparagraph{Estimating $C$:} $C$ is constant matrix that is pre-cached by estimating the uncentered covariance of $i$ from a sample of $100K$ LAION image captions, $C^{-1}$ is its inverse, and both are positive definite (and symmetric).

\myparagraph{Global Key Locking:}
Here we describe our inference time approach called \textit{global} key-locking.

\begin{enumerate}
    \item Given a key-locked trained concept and a prompt that includes it, we start by making a forward pass through the text encoder and calculate the K and V sequence activations that this prompt elicits, denoted by $K$, and $V$. %
    \item We replace the concept word in the prompt by its superclass word and make another forward pass through the text encoder.  This time, we only calculate the K sequence activations that the super-class prompt elicits, denoted by $\widetilde{K}$.
    \item We override $K$ with $\widetilde{K}$ and proceed to sample the image as usual.
\end{enumerate}

We refer to this variant as \textit{global key-locking}, because it locks the K pathway of the entire prompt.

\section{Pseudo code}
\label{supp_pseudo_code}

Algorithm \ref{algobox} shows a pseudo code for the Rank-One Edit Neural Module. This module implements a single-layer edit and the same implementation is used for all layers. It is presented in a PyTorch module style for convenience. The module replaces the keys (K) and values (V) projection modules in the cross-attention layers during training time. It implements \eqref{eq_gated_single} jointly with an online estimation of $i_*$. The only optimized variable is $\ostar$, when $o_*\text{.require\_grad}$ is $True$. For inference with a single concept, we can simply disable the online estimation of $i_*$.

\begin{algorithm}[h]
\SetKwInOut{Input}{Input}
\SetKwInOut{Output}{Output}
\SetKwFunction{init}{init}
\SetKwFunction{forward}{forward}

\BlankLine
\BlankLine
\BlankLine

\textbf{Init Constants:} $W, C^{-1}, \beta, \tau$,

\SetKwProg{Fn}{def}{:}{}
\Fn{\init{\text{init\_input}, \text{is\_key\_locked}, \text{concept\_token\_idx}}}{
	$i_*$ = init\_input[concept\_token\_idx] %

	$o_*$ = W @ init\_input \;

\textit{\\ \# Only train $o_*$ of V pathway} \\

        $o_*\text{.require\_grad}$ = !is\_key\_locked
}

\BlankLine
\BlankLine
\BlankLine

\Fn{\forward{\text{input}, \text{concept\_token\_idx}}}{
\textit{\# online estimation of $i_*$:} \\

	concept\_encoding = input[concept\_token\_idx]
	$i_* = 0.99 * i_* + 0.01 * \text{concept\_encoding}$

 \textit{\\ \# Eq. \ref{eq_gated_single} components:} \\

	$i_*\text{\_energy} = (C^{-1} @ i_*)^T @ i_*$

	$\text{sim} = (\text{input} @ (C^{-1} @ i_*)^T)$

	$\text{sigmoid\_term} = \sigma \left( \frac{(\text{sim} / i_*\text{\_energy}) - \beta}{\tau} \right) $

      \textit{\# calculating $W \Em^\perp$:} \\
	$\text{W\_em\_orthogonal\_term} =  W @ \text{input} - (\text{sim} \cdot (W @ i_*) / i_*\text{\_energy})$

        \textit{\# Eq. \ref{eq_gated_single}:} \\
 	$h = \text{W\_em\_orthogonal\_term} + \text{sigmoid\_term} @ o_* $

        \KwRet{$h$}
}

\caption{Rank-One Edit Module for a Single Concept (Eq. \ref{eq_gated_single})}\label{algobox}
\end{algorithm}

Where ``$@$'' operator denotes matrix multiplication, $init\_input$ is the encoding
 of a prompt saying ``A photo of a <superclass\_word>'', $input$ is the encoding $\ve$ of an entire prompt,
$W$ denotes the pretrained layer weights, $concept\_token\_idx$ is the index of the concept word $S^*$, $C^{-1}$ is the inverse of the constant matrix $C$, and $\beta, \tau$ are the sigmoid hyper-parameters.
Keep in mind that because \eqref{eq_gated_single} is valid for all $m$, this module processes the entire prompt simultaneously.

\section{Training classes}
\begin{itemize}
    \item cat toy (from Textual-inversion)
    \item headless scuplture  (from Textual-inversion)
    \item cat (from Custom-Diffusion)
    \item chair (from Custom-Diffusion)
    \item wooden pot (from Custom-Diffusion)
    \item dog (from Custom-Diffusion)
    \item teaddybear (from Custom-Diffusion)
    \item tortoise plushy (from Custom-Diffusion)
    \item puppy (from DreamBooth)
    \item sunglasses (from DreamBooth)
    \item teapot (similar to DreamBooth teapot)
\end{itemize}

\section{Success and Failures of Global Locking}
\label{sec_suppl_global_head_tail}
\begin{figure}[h]
\includegraphics[width=\columnwidth, trim={0.cm 0cm 0cm 0cm},clip]{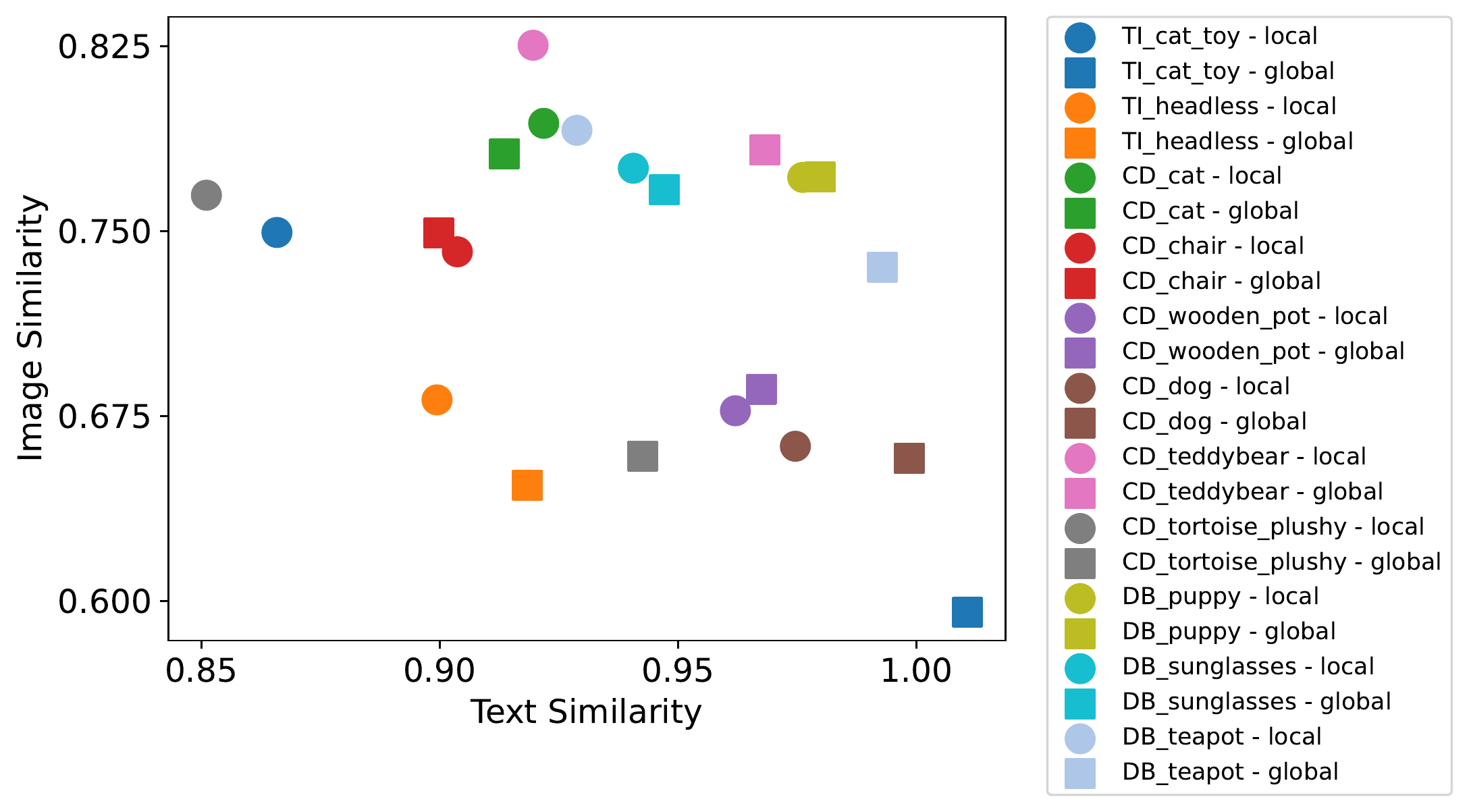} %
\caption{ \textbf{Success and failures of global locking:} Circles - using  local locking. Square - using global locking. Global locking struggles with classes of uniquely shaped object like CD\_tortoise\_plushy. However it is successful with classes of everyday concepts like CD\_dog.
}
\label{fig_global_local_scatter}
\vspace{-5pt}
\end{figure}

Image-similarity of global locking may appear worse than local locking, but it is not always the case. We find that global locking struggles with uniquely shaped objects, but is successful with common objects like pets, teddy bears, and chairs. \figref{fig_global_local_scatter} shows how global locking struggles with uniquely shaped objects, but is successful with everyday concepts. In cases where global locking is successful, we find that it allows to generate rich scenes and portrays better the nuances of the object attributes or activities.

\section{Training prompts}
We used the following set of prompts for training:
\begin{itemize}
    \item a photo of a $S^*$
    \item a good photo of a $S^*$
    \item the photo of a $S^*$
    \item a good photo of the $S^*$
    \item image of a $S^*$
    \item image of the $S^*$
    \item A photograph of $S^*$
    \item A $S^*$ shown in a photo,
    \item A photo of $S^*$
\end{itemize}

\section{Qualitative figures parameters choice}
We describe the inference-time parameters we chose for each prompt in qualitative figures:

\myparagraph{Figure \ref{fig1}:}
\begin{itemize}
    \item "A Teapot* made out of \{pure gold, yarn, glass\}" = Lock: Global, Train Bias: 0.75, Infer Bias: 0.45, Infer temp: 0.1
    \item "A Teapot* oil painting ghibli inspired" = Lock: Global, Train Bias: 0.75, Infer Bias: 0.45, Infer temp: 0.1
    \item "A Teddy* dressed as a \{samurai, superhero, wizard\}" = Lock: Global, Train Bias: 0.75, Infer Bias: 0.45, Infer temp: 0.1
    \item "A Teddy* dressed in a blue suit is cooking a gourmet meal" = Lock: local, Train Bias: 0.75, Infer Bias: 0.75, Infer temp: 0.1
    \item "A Teddy* is sailing inside a Teapot* in a lake" = Lock: local, Train Bias: 0.75, Infer Bias: 0.55, Infer temp: 0.15
    \item "A Teddy* sitting by the fire with a Teapot* on a chilly night" = Lock: local, Train Bias: 0.75, Infer Bias: 0.55, Infer temp: 0.15
    \item "Painting of a Teddy* is sitting next to a Teapot* on a picnic" = Lock: local, Train Bias: 0.75, Infer Bias: 0.55, Infer temp: 0.15
    \item "The Teddy* in engraved on a Teapot*" = Lock: local, Train Bias: 0.75, Infer Bias: 0.55, Infer temp: 0.15
\end{itemize}

\myparagraph{Figure \ref{fig_single}:}
\begin{itemize}
    \item "A Sculpture* wearing a sombrero" = Lock: local, Train Bias: 0.55, Infer Bias: 0.65, Infer temp: 0.15
    \item "A Cat* is acting in a play wearing a costume" = Lock: global, Train Bias: 0.75, Infer Bias: 0.65, Infer temp: 0.15
    \item "A Teddy* is playing with a ball in the water" = Lock: local, Train Bias: 0.75, Infer Bias: 0.75, Infer temp: 0.1
    \item "A broken Pot*" = Lock: global, Train Bias: 0.75, Infer Bias: 0.45, Infer temp: 0.15
    \item "A Dog* is reading a book" = Lock: global, Train Bias: 0.75, Infer Bias: 0.45, Infer temp: 0.15
\end{itemize}

\myparagraph{Figure \ref{fig_multi}:}
\begin{itemize}
    \item "Watercolor painting of Cat* sitting on Chair*" = Lock: local, Train Bias: 0.6, Infer Bias: 0.6, Infer temp: 0.1
    \item "Photo of a Table* and the Chair*" = Lock: local, Train Bias: 0.7, Infer Bias: 0.75, Infer temp: 0.1
    \item "The Cat* playing with a Pot* in a garden" = Lock: local, Train Bias: 0.75, Infer Bias: 0.4, Infer temp: 0.1
    \item "A Teddy* is sitting in the garden and wearing the Sunglasses*" = Lock: local, Train Bias: 0.75, Infer Bias: 0.55, Infer temp: 0.15
\end{itemize}

\section{User Study Details}
\label{suppl_user_study}
We evaluate the models by two user studies conducted with Amazon Mechanical Turk. In the first, raters were asked to rank a set of images generated by the three methods (\ourmethod, CD, DB). In the second, we compared \ourmethod to ``vanilla'' stable-diffusion (SD) in order to study whether \ourmethod harms the generative prior.

\subsection{Personalization Method Comparison}
\begin{figure*}[h]
\includegraphics[width=0.9\textwidth, trim={0.cm 20.cm 0.cm 0.cm},clip]{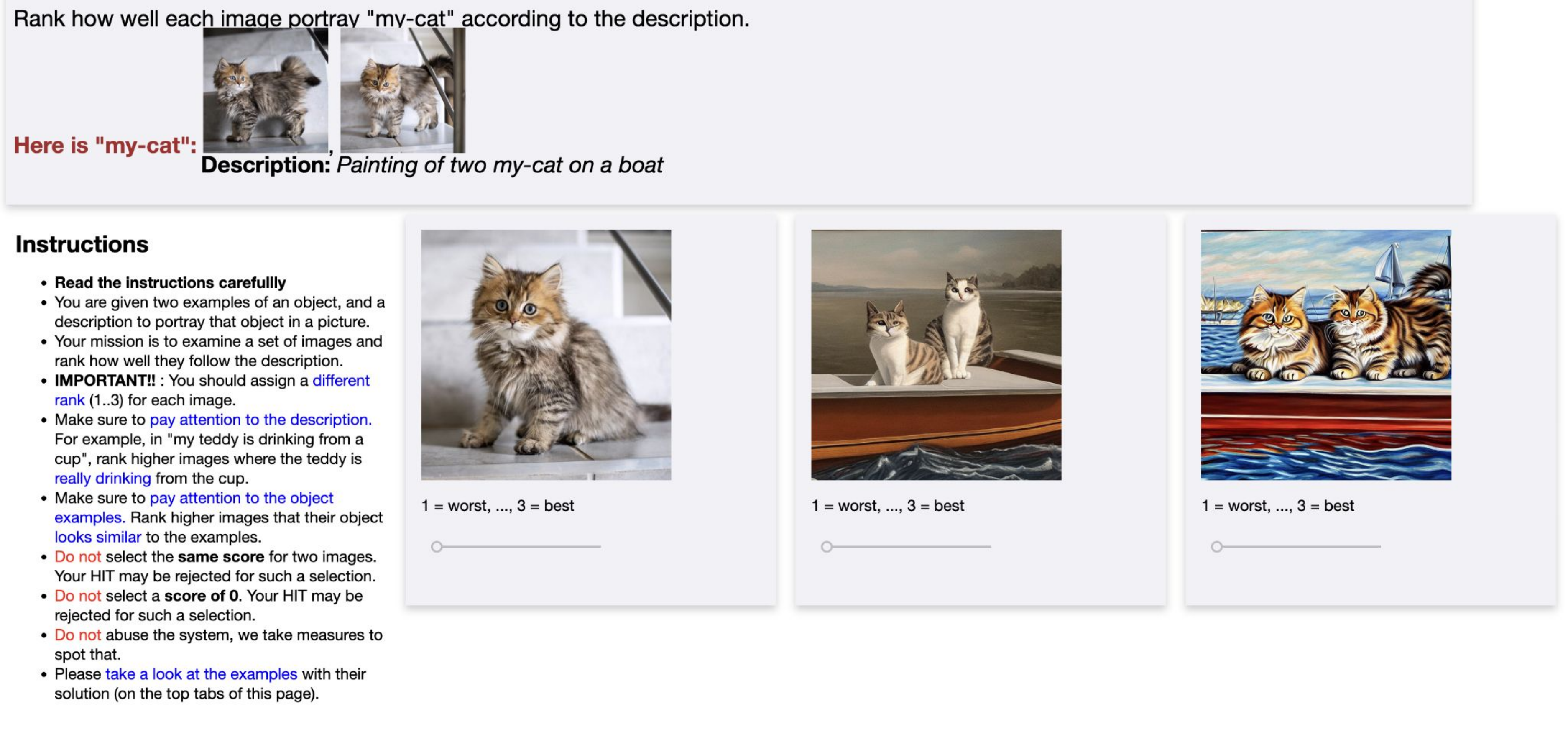} %
\vspace{-15pt}
\caption{\edit{One qualification trial of the method comparison user study. This was the hardest for the prescreened pool of raters, and we made sure all qualified users were tested with it.} }
\label{fig_amt1_task}

\begin{flushleft}
\includegraphics[width=0.7\textwidth, trim={0.cm 0.cm 0.cm 0.cm},clip]{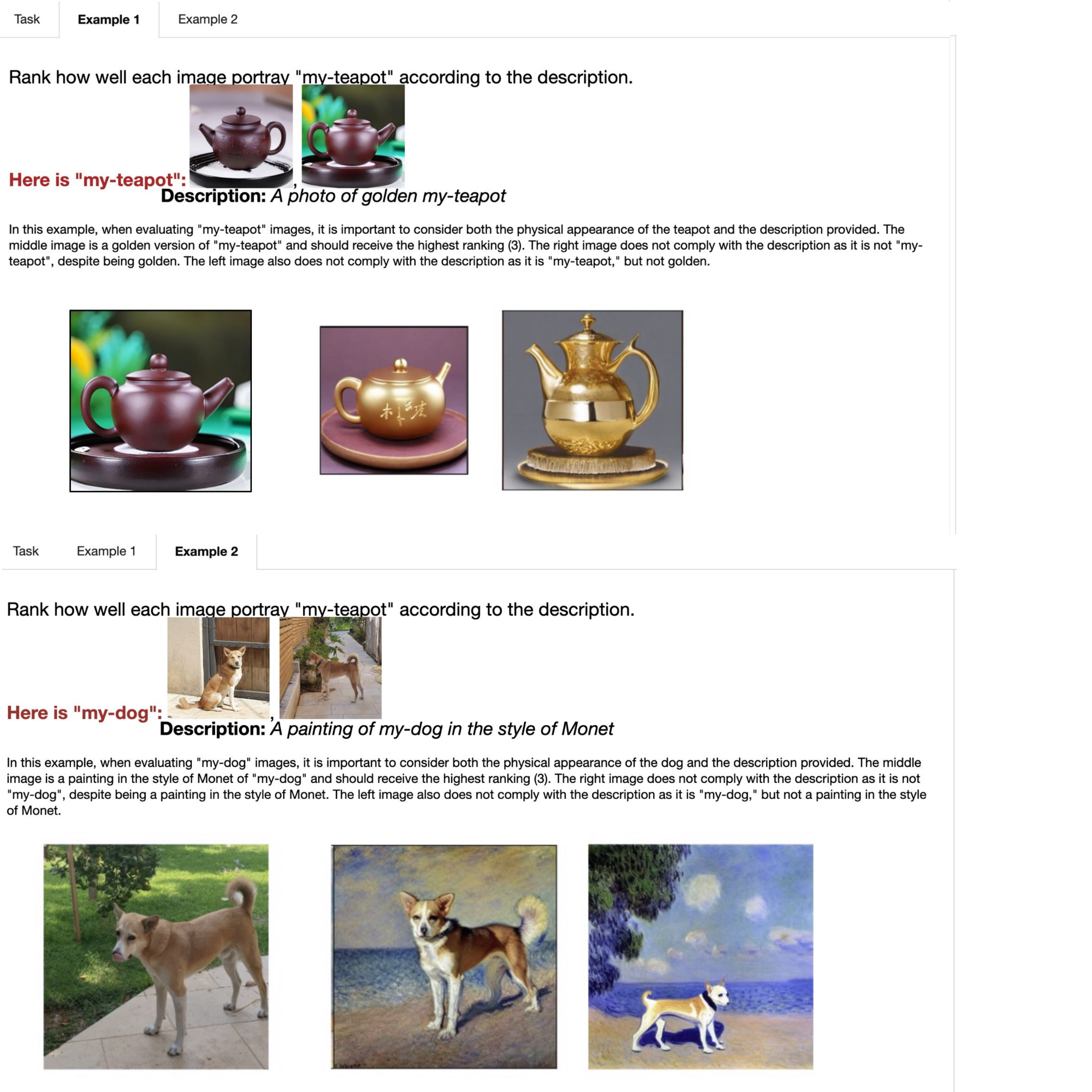} %
\caption{\edit{Examples provided in the method comparison user study. } }
\label{fig_amt1_examples}
\end{flushleft}
\end{figure*}

For the first study, in each trial, raters were shown two train set images of a concept and a prompt description. They were then asked to examine a set of three generated images, one from each method, and assign each image a unique rank from 1 to 3 (best), based on how well they portrayed the concept according to the description. Method order was randomized in each trial.
\figref{fig_amt1_task} illustrates the experimental framework that was used in trials. \figref{fig_amt1_examples} displays the examples provided to help guide raters through the instructions.

We had $2104$ trials from $11$ concepts, with an average of $24$ prompts per concept, and $8$ trials per prompt. We used all the prompts from the challenging subset of \textit{per-group} prompts.
For \ourmethod method images, we select the runtime variant with the highest harmonic mean for text and visual similarities in each \textit{prompt}.  Nonetheless, both methods require the same computational resources. Reported rank scores are balanced ``per-class'' \cite{DRAGON}. Namely, we compute the mean score per concept class, and then uniformly average between all class scores.

We paid \$0.12 per trial. To maintain the quality of the queries, we only picked raters with AMT ``masters'' qualification, demonstrating a high degree of approval rate over a wide range of tasks. Furthermore, we also conducted a qualification test on the prescreened pool of raters, consisting of a few curated trials that were very simple. In each qualification trial, we included one well generated image that depict the concept as described by the prompt, one image generated from vanilla stable-diffusion, and one image from the concept training examples. We only qualified raters who had completed a minimum of 5 trials out of a pool of 11 trials with perfect scores. The reason for not having all raters complete all trials is that AMT assigns them randomly. In one of our qualification trials, which is demonstrated in \figref{fig_amt1_task}, a significant proportion of raters in the prescreened pool did not perform well. As a result, we replicated the task and made sure that all qualified raters had successfully completed it.

\subsection{Generative Prior Preservation}
\begin{figure*}[h]
\includegraphics[width=\textwidth, trim={0.cm 22.cm 0.cm 0.cm},clip]{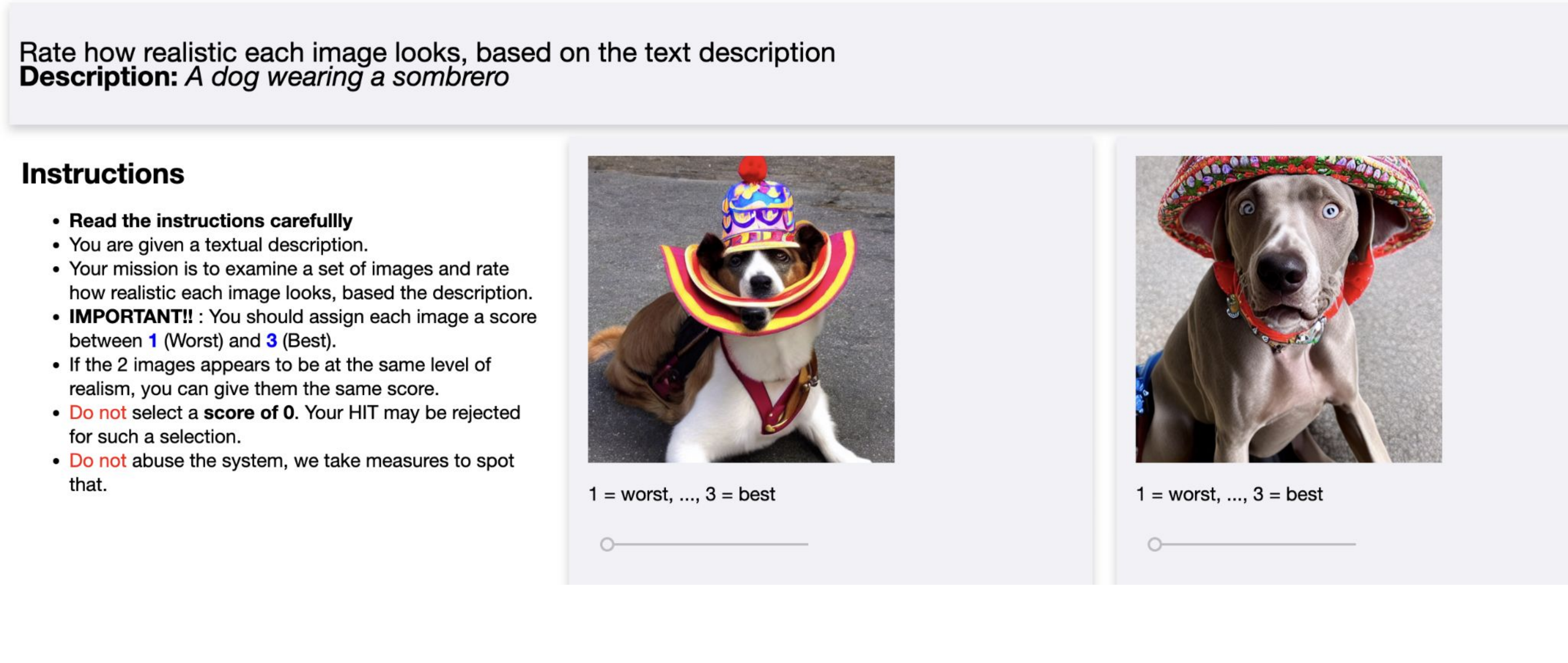} %
\vspace{-15pt}
\caption{\edit{One trial of the prior preservation user study. } }
\label{fig_amt2_task}

\begin{flushleft}
\includegraphics[width=0.8\textwidth, trim={0.cm 3.cm 0.cm 0.cm},clip]{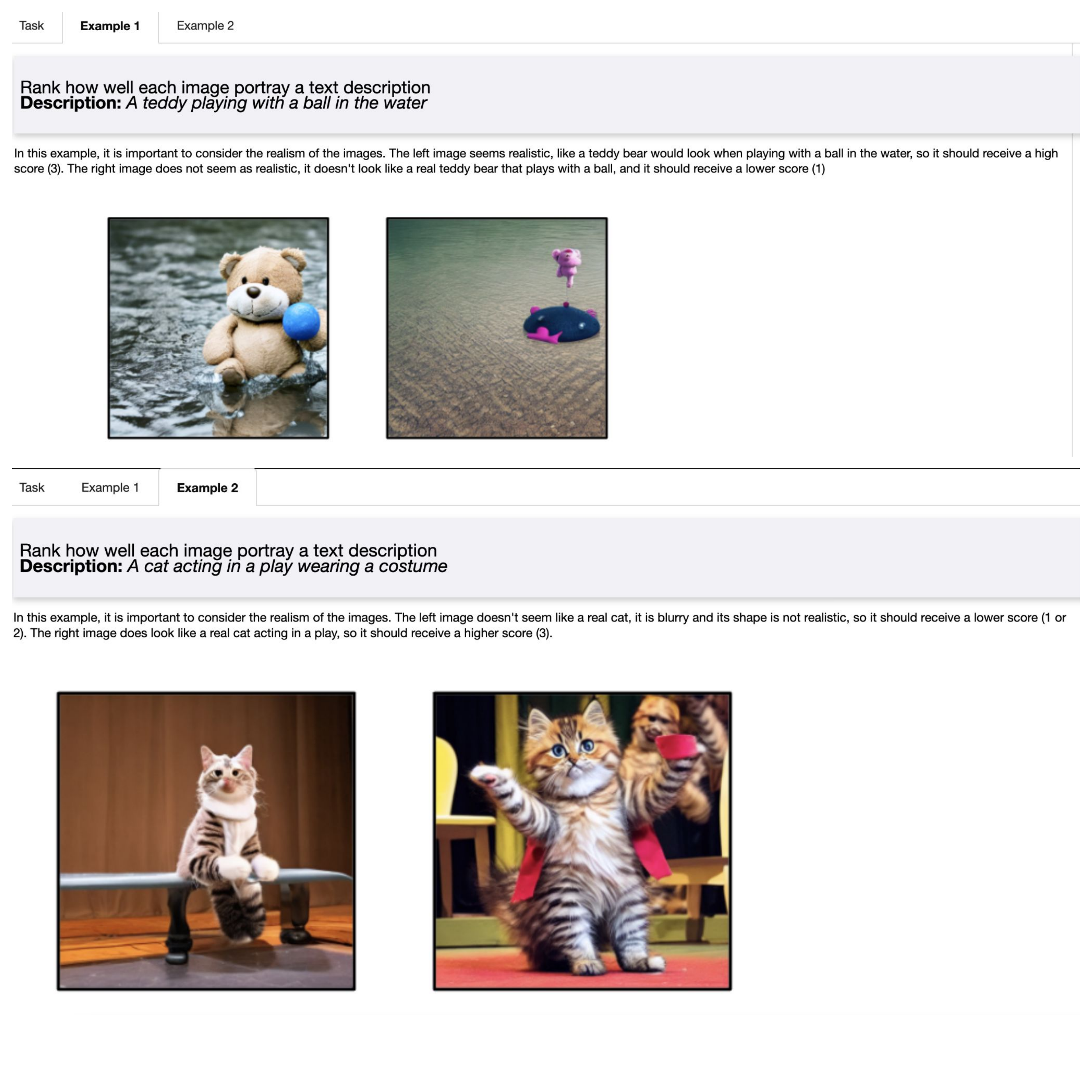} %
\vspace{-10pt}
\caption{\edit{Examples provided in the prior preservation user study. } }
\label{fig_amt2_examples}
\end{flushleft}
\end{figure*}

For the second study, in each trial, raters were given a prompt description and two generated images, one by \ourmethod and another by SD.  They were asked to rate how realistic each image looks, based on the text description. The participants were instructed to assign each image a score between 1 to 3 (best). Unlike the first study, the raters were allowed to give the same score for both images. Method order was randomized in each trial. We had the same number of trials, prompts and concepts as in the first study. Additionally, we utilized the same images generated for \ourmethod.
\figref{fig_amt2_task} illustrates the experimental framework that was used in trials. \figref{fig_amt2_examples} displays the examples provided to help guide raters through the instructions.

We paid \$0.1 per trial. To maintain the quality of the queries, we only picked raters with AMT ``masters'' qualification. Furthermore, we also executed a qualification test with a few curated trials that are very simple. We only qualified raters who had completed all trials with perfect scores and had completed at least 5 trials.

\end{document}